\documentclass[10pt,logo,copyright]{nvidiatechreport}
\usepackage[numbers,sort&compress]{natbib}

\usepackage[utf8]{inputenc} %
\usepackage[T1]{fontenc}    %

\usepackage{parskip}        %
\usepackage{url}            %
\usepackage{booktabs}       %
\usepackage{amsfonts}       %
\usepackage{nicefrac}       %
\usepackage{microtype}      %
\usepackage{xcolor}         %
\usepackage[dvipsnames]{xcolor} %
\usepackage{graphicx}
\usepackage{animate}        %
\usepackage{subcaption}
\usepackage{tabularx}
\usepackage{makecell}
\usepackage{adjustbox}
\usepackage{setspace}
\newcolumntype{M}[1]{>{\centering\arraybackslash}m{#1}}
\usepackage{float}
\usepackage{tikz}
\usetikzlibrary{positioning,shapes,arrows}
\usepackage{amsmath,amsfonts,bm, bbm,leftindex}
\usepackage{multirow}
\usepackage{comment}
\usepackage{gensymb}
\usepackage{lipsum}
\usetikzlibrary{arrows.meta, positioning, fit}
\usepackage[para]{threeparttable}
\usepackage{tikz}
\usetikzlibrary{tikzmark}
\usepackage{float}
\usepackage{listings}
\usepackage{xcolor}
\usepackage{adjustbox}
\usepackage{multirow}
\usepackage{makecell}

\def\eqref#1{equation~\ref{#1}}

\def\1{\bm{1}}

\DeclareMathAlphabet{\mathsfit}{\encodingdefault}{\sfdefault}{m}{sl}
\SetMathAlphabet{\mathsfit}{bold}{\encodingdefault}{\sfdefault}{bx}{n}

\makeatletter
\let\save@mathaccent\mathaccent
\newcommand*\if@single[3]{%
  \setbox0\hbox{${\mathaccent"0362{#1}}^H$}%
  \setbox2\hbox{${\mathaccent"0362{\kern0pt#1}}^H$}%
  \ifdim\ht0=\ht2 #3\else #2\fi
  }
\newcommand*\rel@kern[1]{\kern#1\dimexpr\macc@kerna}
\newcommand*\widebar[1]{\@ifnextchar^{{\wide@bar{#1}{0}}}{\wide@bar{#1}{1}}}
\newcommand*\wide@bar[2]{\if@single{#1}{\wide@bar@{#1}{#2}{1}}{\wide@bar@{#1}{#2}{2}}}
\newcommand*\wide@bar@[3]{%
  \begingroup
  \def\mathaccent##1##2{%
    \let\mathaccent\save@mathaccent
    \if#32 \let\macc@nucleus\first@char \fi
    \setbox\z@\hbox{$\macc@style{\macc@nucleus}_{}$}%
    \setbox\tw@\hbox{$\macc@style{\macc@nucleus}{}_{}$}%
    \dimen@\wd\tw@
    \advance\dimen@-\wd\z@
    \divide\dimen@ 3
    \@tempdima\wd\tw@
    \advance\@tempdima-\scriptspace
    \divide\@tempdima 10
    \advance\dimen@-\@tempdima
    \ifdim\dimen@>\z@ \dimen@0pt\fi
    \rel@kern{0.6}\kern-\dimen@
    \if#31
      \overline{\rel@kern{-0.6}\kern\dimen@\macc@nucleus\rel@kern{0.4}\kern\dimen@}%
      \advance\dimen@0.4\dimexpr\macc@kerna
      \let\final@kern#2%
      \ifdim\dimen@<\z@ \let\final@kern1\fi
      \if\final@kern1 \kern-\dimen@\fi
    \else
      \overline{\rel@kern{-0.6}\kern\dimen@#1}%
    \fi
  }%
  \macc@depth\@ne
  \let\math@bgroup\@empty \let\math@egroup\macc@set@skewchar
  \mathsurround\z@ \frozen@everymath{\mathgroup\macc@group\relax}%
  \macc@set@skewchar\relax
  \let\mathaccentV\macc@nested@a
  \if#31
    \macc@nested@a\relax111{#1}%
  \else
    \def\gobble@till@marker##1\endmarker{}%
    \futurelet\first@char\gobble@till@marker#1\endmarker
    \ifcat\noexpand\first@char A\else
      \def\first@char{}%
    \fi
    \macc@nested@a\relax111{\first@char}%
  \fi
  \endgroup
}
\makeatother

\definecolor{darkred}{rgb}{0.7, 0.0, 0.0}

\usepackage{pifont}

\usepackage[nameinlink]{cleveref}
\crefname{equation}{Eq.}{Eqs.}
\crefname{figure}{Fig.}{Figs.}
\crefname{section}{Sec.}{Sec.}
\crefname{appendix}{App.}{App.}
\crefname{table}{Tab.}{Tabs.}
\crefname{algorithm}{Algo}{Algo}
\crefname{thm}{Thm}{Thm}
\Crefname{thm}{Thm}{Thm}
\crefname{prop}{Prop}{Prop}

\newcommand{\crefnames}[3]{%
  \@for\next:=#1\do{%
    \expandafter\crefname\expandafter{\next}{#2}{#3}%
  }%
}

\title{Describe Anything: Detailed Localized Image and Video Captioning}

\author{Long Lian\textsuperscript{1,2} \quad Yifan Ding\textsuperscript{1} \quad Yunhao Ge\textsuperscript{1} \quad Sifei Liu\textsuperscript{1} \quad Hanzi Mao\textsuperscript{1} \quad Boyi Li\textsuperscript{1,2} \quad Marco Pavone\textsuperscript{1} \quad Ming-Yu~Liu\textsuperscript{1} \quad Trevor Darrell\textsuperscript{2} \quad Adam Yala\textsuperscript{2,3} \quad Yin Cui\textsuperscript{1} \\
\textsuperscript{1}NVIDIA \quad \textsuperscript{2}UC Berkeley \quad \textsuperscript{3}UCSF \\
}

\renewcommand{\today}{}

\newcommand{\ModelName}{DAM}

\begin{document}

\def\figDemo{
\twocolumn[{
\renewcommand\twocolumn[1][]{##1}
\maketitle
\begin{center}
  \captionsetup{type=figure}
  \includegraphics[width=0.9\linewidth]{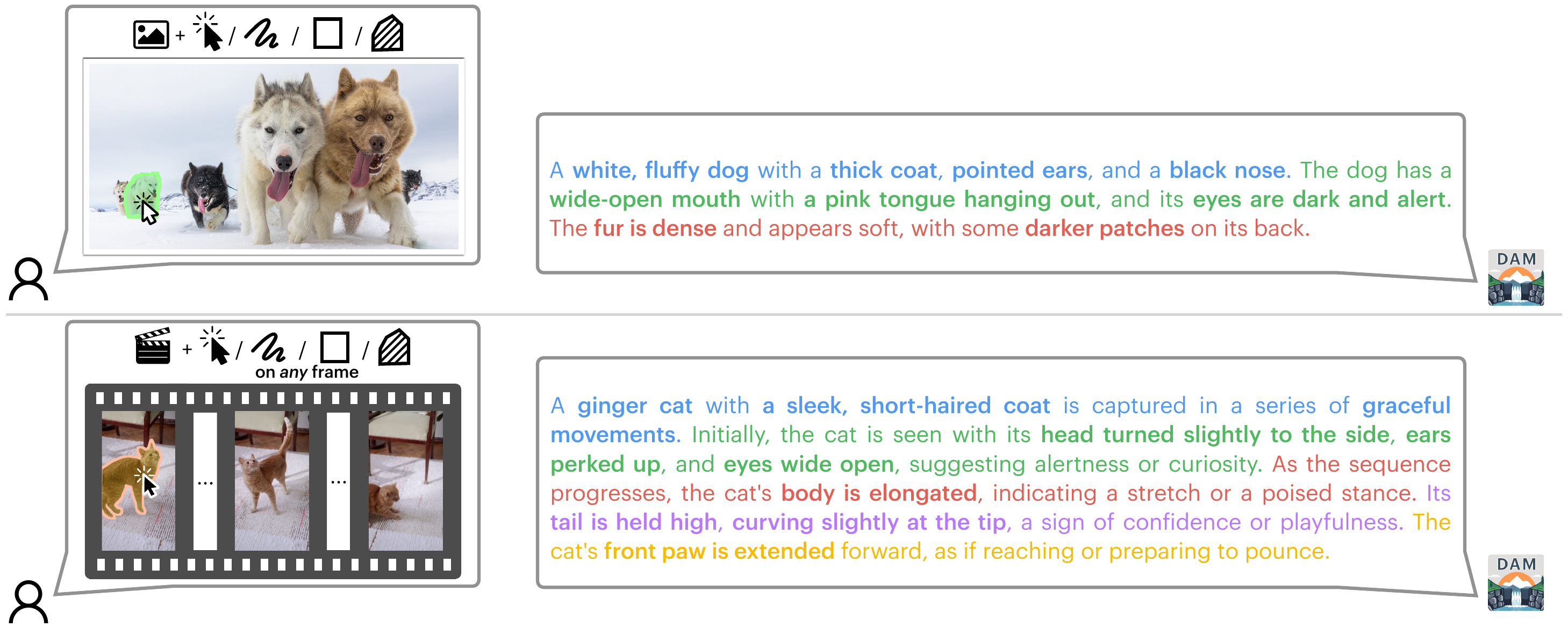}
  \captionof{figure}{
    \textbf{Describe Anything Model (\ModelName{})} generates \textbf{detailed localized captions} for user-specified regions within \textbf{images}~(top) and \textbf{videos}~(bottom). DAM accepts various region specifications, including clicks, scribbles, boxes, and masks. For videos, specifying the region in \textit{any frame} suffices. \vspace{8mm}
  }
  \label{fig:demo}
\end{center}
}]
}

\def\figContradiction#1{
\begin{figure*}[#1]
\centering

\begin{subfigure}{0.95\linewidth}
    \centering
    \includegraphics[width=\linewidth]{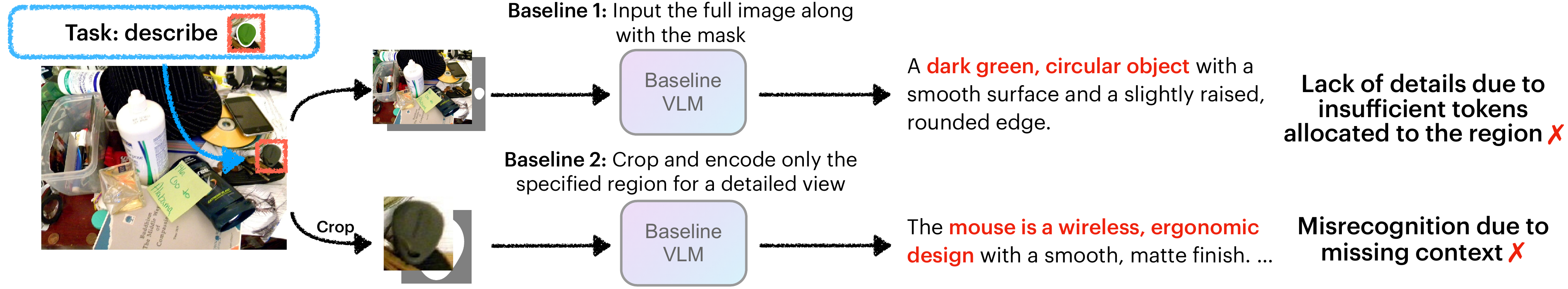}
    \caption{Baseline approaches using existing VLMs struggle to balance details for the region with contextual information.}
    \label{fig:contradiction_a}
\end{subfigure}
\vspace{4mm}
\begin{subfigure}{0.95\linewidth}
    \centering
    \includegraphics[width=\linewidth]{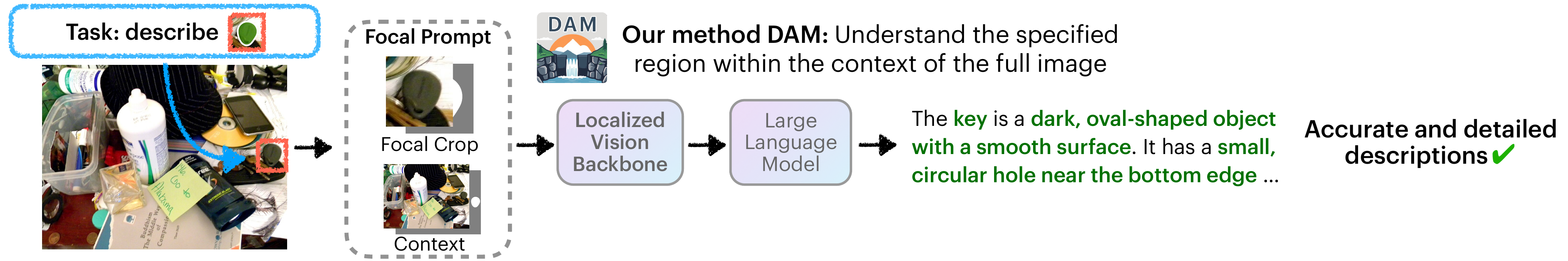}
    \caption{Our method DAM resolves this contradiction by understanding the user-specified region within its broader context.}
    \label{fig:contradiction_b}
\end{subfigure}
\vspace{-2mm}
\caption{Existing VLMs struggle to generate accurate and detailed descriptions for regions due to a trade-off between regional details and contextual information, shown in \textbf{(a)}. Our method, shown in \textbf{(b)}, perceives regions within their context, ensuring precise descriptions by balancing our focus with contextual cues through \textit{focal prompt} and \textit{localized vision backbone}. Notably, our method does \textit{not} increase the number of image tokens compared to encoding only the full image, maintaining efficiency.\vspace{-3.5mm}}
\label{fig:contradiction}
\end{figure*}
}

\def\figArchComparison#1{
\begin{figure}[#1]
\centering
\includegraphics[width=1.0\linewidth]{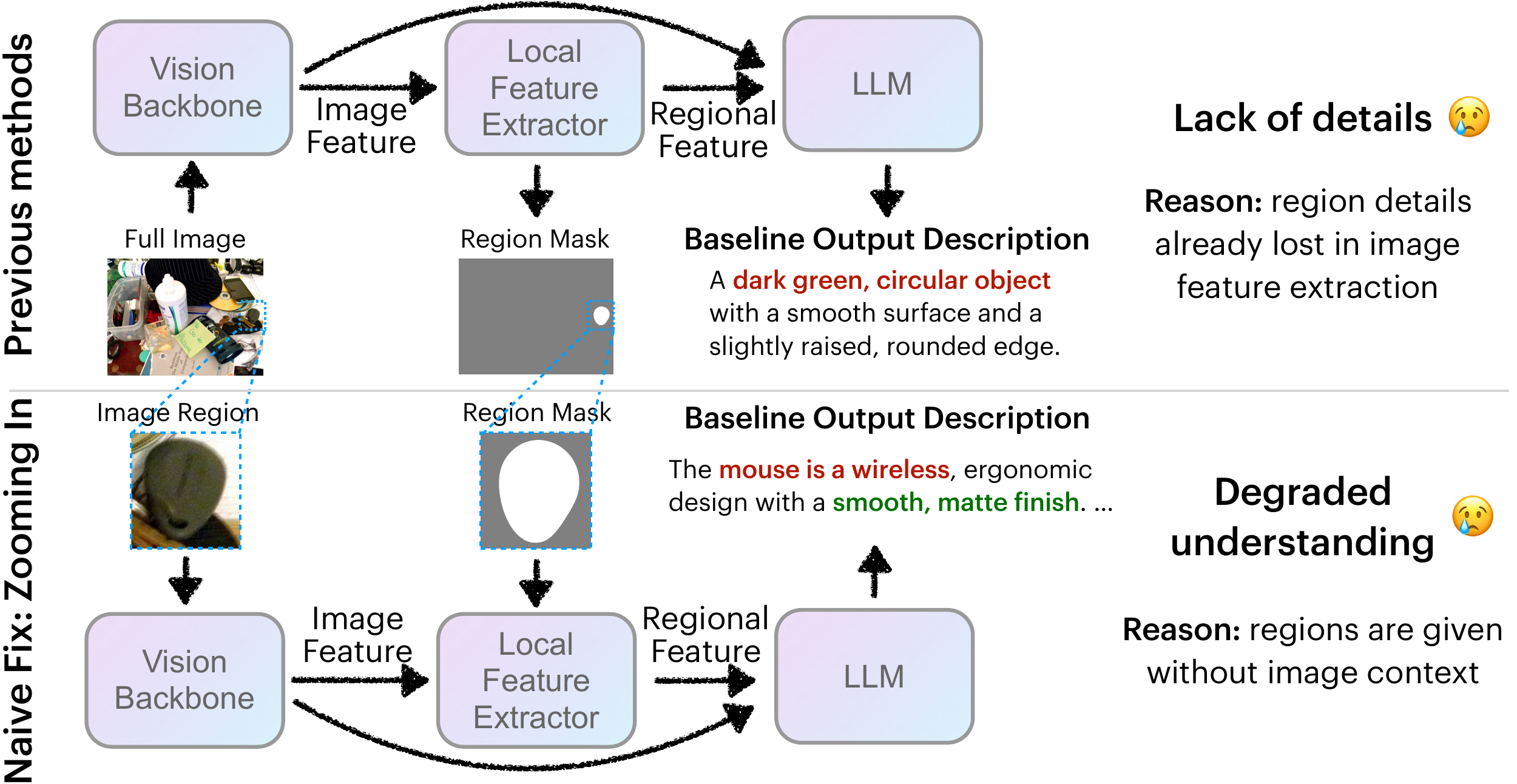}
\vspace{-4mm}
\caption{\textbf{Top:} Prior regional captioners derive regional features from global image representations, leading to vague descriptions.  
\textbf{Bottom:} Zooming in (cropping the image region) enhances detail but loses contextual cues, degrading recognition. This underscores the need for a design that \textbf{encodes detail-rich regional features while preserving context for improved DLC performance}.\vspace{-4mm}}
\label{fig:arch_comparison}
\end{figure}
}

\def\figArchitecture#1{
\begin{figure}[#1]
\centering
\includegraphics[width=0.8\linewidth]{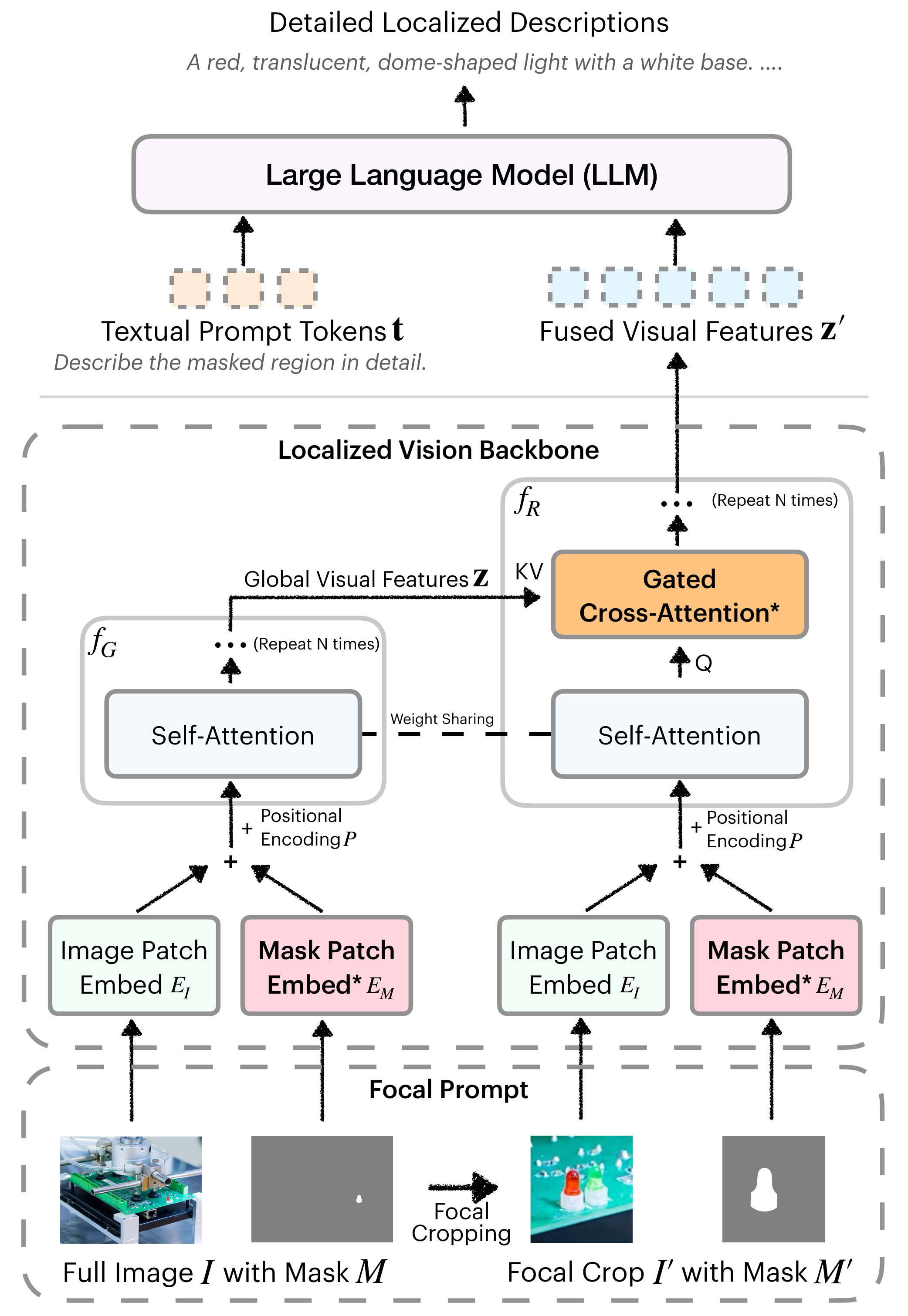}
\caption{\textbf{Architecture of the Describe Anything Model (DAM).} DAM employs a \textit{focal prompt} to encode user-specified regions with high token density while preserving context for detailed understanding. Focal cropping is applied to the image and its corresponding mask, retaining surrounding areas for local context. Both the full image and the focal crop are the inputs into the \textit{localized vision backbone}, where images and binary masks are embedded in a spatially aligned fashion. Global context from the full image is leveraged to help understand the focal crop through gated cross-attention. The resulting visual features and prompt tokens are fed into a large language model to generate detailed, context-aware descriptions. * indicates initialized to output zeros.}
\label{fig:architecture}
\end{figure}
}

\def\figBenchmark#1{
\begin{figure*}[#1]
\centering
\includegraphics[width=0.95\linewidth]{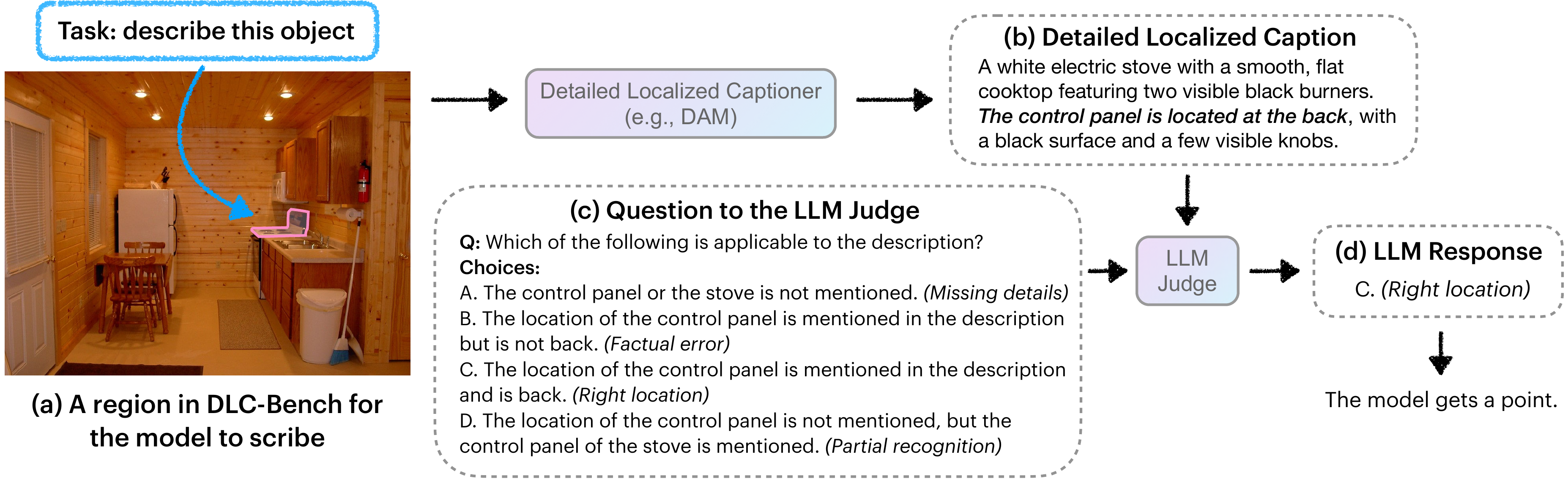}
\vspace{-2mm}
\caption{\textbf{We propose DLC-Bench, a benchmark tailored to detailed localized captioning.} In DLC-Bench, a captioning model is prompted to describe a specified image region (a). The generated description (b) is then evaluated by querying an LLM Judge (c). Points are assigned or deducted based on the LLM's response (d). The question we show in (c) is an example of positive questions.}
\label{fig:benchmark}
\end{figure*}
}

\def\tabMainResult#1{
\begin{table}[#1]
    \centering
    \setlength{\tabcolsep}{1.5pt}
    \begin{adjustbox}{width=0.9\linewidth,center}
    \begin{tabular}{lc@{\hspace{2.5mm}}c@{\hspace{3.0mm}}c@{\hspace{3.0mm}}c}
        \toprule
        {\small \textbf{Method}} & {\small \hspace{-5mm}\textbf{\#Params}} & {\small \textbf{Pos (\%)}} & {\small \textbf{Neg (\%)}} & {\small \textbf{Avg (\%)}} \\
        \midrule
        \multicolumn{5}{l}{\textit{General VLMs:}} \\
        GPT-4o~\cite{gpt4o} & - & 43.4 & \underline{79.6} & 61.5 \\
        o1~\cite{o1}$^\dagger$ & - & \underline{46.3} & 78.8 & \underline{62.5} \\
        Claude 3.7 Sonnet~\cite{team2025claude}$^\dagger$ & - & 21.8 & 50.4 & 36.1 \\
        Gemini 2.5 Pro~\cite{team2023gemini,team2025gemini}$^\dagger$ & - & 36.5 & 75.2 & 55.8 \\
        Llama-3.2 Vision~\cite{dubey2024llama}\hspace{-5mm} & 11B & 30.7 & 63.8 & 47.3 \\
        VILA1.5-Llama-3~\cite{lin2024vila}\hspace{-5mm} & 8B & 22.5 & 61.0 & 41.8 \\
        InternVL2.5~\cite{chen2023internvl,chen2024expanding,wang2024mpo} & 8B & 15.9 & 42.0 & 28.9 \\
        LLaVA v1.6~\cite{liu2024visual,liu2023improvedllava,liu2024llavanext} & 7B & 15.4 & 55.0 & 35.2 \\
        Qwen2.5-VL~\cite{wang2024qwen2,Qwen2.5-VL} & 7B & 20.3 & 62.2 & 41.2 \\
        VILA1.5~\cite{lin2024vila} & 3B & 16.0 & 50.0 & 33.0 \\
        \midrule
        \multicolumn{5}{l}{\textit{Region-specific VLMs (full / cropped input):}} \\
        GPT4RoI~\cite{zhang2023gpt4roi} & 7B & 6.5/3.5 & 46.2/52.0 & 26.3/27.7 \\
        Shikra~\cite{chen2023shikra} & 7B & 2.7/8.0 & 41.8/51.4 & 22.2/29.7 \\
        Ferret~\cite{you2023ferret} & 7B & 6.4/14.2 & 38.4/46.8 & 22.4/30.5 \\
        RegionGPT~\cite{guo2024regiongpt} & 7B & 13.0/10.6 & 41.4/46.4 & 27.2/28.5 \\
        ControlCap~\cite{zhao2024controllable} & 0.3B & 18.3/~~3.6 & 75.6/53.6 & 47.0/28.6 \\
        SCA~\cite{huang2024segment} & 3B & ~~3.4/~~0.1 & 44.6/18.4 & 24.0/~~9.3 \\
        OMG-LLaVA~\cite{zhang2024omg} & 7B & ~~0.9/~~5.6 & 16.0/32.6 & ~~8.5/19.1 \\
        VP-SPHINX~\cite{lin2024draw} & 13B & 11.7/26.3 & 33.2/71.6 & 22.5/49.0 \\
        \midrule
        \textbf{DAM (Ours)} & 3B & \textbf{52.3} & \textbf{82.2} & \textbf{67.3} \\
        \bottomrule
    \end{tabular}
    \end{adjustbox}
    \vspace{-2mm}
    \caption{\textbf{Accuracies on detailed localized captioning in our proposed DLC-Bench.} DAM outperforms previous API-only models, open-source models, and region-specific VLMs on detailed localized captioning. \underline{Underlined}: the second-best method. $\dagger$: models with thinking mode.
    }
    \label{tab:main_result}
\end{table}
}

\def\tabAblationResult#1{
\begin{table}[#1]
\centering
\setlength{\tabcolsep}{1.5pt}
\begin{adjustbox}{width=0.8\linewidth,center}
\begin{tabular}{l@{\hspace{-2pt}}ccccc@{\hspace{-4pt}}l}
\toprule
{\small \textbf{Prompting}} & {\small \textbf{XAttn}} & {\small \textbf{\#IT}} & {\small \textbf{Pos (\%)}} & {\small \textbf{Neg (\%)}} & {\small \textbf{Avg (\%)}} & \\
\midrule
{\small Full Image Only}       & {\small No} & 196 & 32.1  & 65.4  & 48.7 & \\
{\small Local Crop Only}       & {\small No} & 196 & 43.5  & 76.6  & 60.1 &{\fontsize{8}{10}\selectfont \textbf{\textcolor{Green}{(\textbf{+11.4})}}} \\
{\small Full + Local Crop}     & {\small No*} & 392 & 26.3  & 58.6  & 42.4 &{\fontsize{8}{10}\selectfont \textbf{\textcolor{Red}{(\textbf{-6.3})}}} \\
{\small Full + Local Crop}     & {\small Yes} & 196 & 45.7  & 80.6  & 63.2 &{\fontsize{8}{10}\selectfont \textbf{\textcolor{Green}{(\textbf{+14.5})}}} \\
{\small Focal Crop Only}     & {\small No} & 196 & 47.3  & \textbf{83.6}  & 65.4 &{\fontsize{8}{10}\selectfont \textbf{\textcolor{Green}{(\textbf{+16.7})}}} \\
{\small \textbf{Full + Focal Crop}}   & Yes & 196 & \textbf{52.3}  & 82.2  & \textbf{67.3} &{\fontsize{8}{10}\selectfont \textbf{\textcolor{Green}{(\textbf{+18.6})}}} \\
\bottomrule
\end{tabular}
\end{adjustbox}
\vspace{-2mm}
\caption{\textbf{Ablation studies across different visual prompts, cross-attention settings, and the number of image tokens.} Local crop denotes cropping without surrounding context. XAttn: cross-attention. \#IT: number of image tokens. * indicates the full image and the crop are concatenated on the sequence dimension. \textbf{Bold}: Our proposed \textbf{focal prompt}.}
\label{tab:ablation_result}
\end{table}
}

\def\tabAblationData#1{
\begin{table}[#1]
    \centering
    \setlength{\tabcolsep}{3.5pt}
    \begin{adjustbox}{width=1.0\linewidth,center}
    \begin{tabular}{lcccc@{\hspace{-4pt}}l}
        \toprule
        {\small \textbf{Data}} & \# regions & {\small \textbf{Pos (\%)}} & {\small \textbf{Neg (\%)}} & {\small \textbf{Avg (\%)}} \\
        \midrule
        {\small LVIS} & 373k & 34.0 & 72.6 & 53.3 & \\
        {\small + additional datasets} & 602k & 47.5 & 80.0 & 63.8 &{\fontsize{8}{10}\selectfont \textbf{\textcolor{Green}{(\textbf{+10.5})}}}\\
        {\small + SSL on 10\% of SA-1B} & 1.38M & \textbf{52.3} & \textbf{82.2} & \textbf{67.3} &{\fontsize{8}{10}\selectfont \textbf{\textcolor{Green}{(\textbf{+14.0})}}}\\
        \bottomrule
    \end{tabular}
    \end{adjustbox}
    \vspace{-2mm}
    \caption{\textbf{Our model benefits from diverse datasets generated by DLC-SDP.} Scaling the dataset in size and diversity significantly improves model performance, and SSL further enhances performance using widely available unannotated images.}
\label{tab:ablation_data}
\end{table}
}

\def\figComparison#1{
\begin{figure}[#1]
\centering
\vspace{-2mm}
\includegraphics[width=1.0\linewidth]{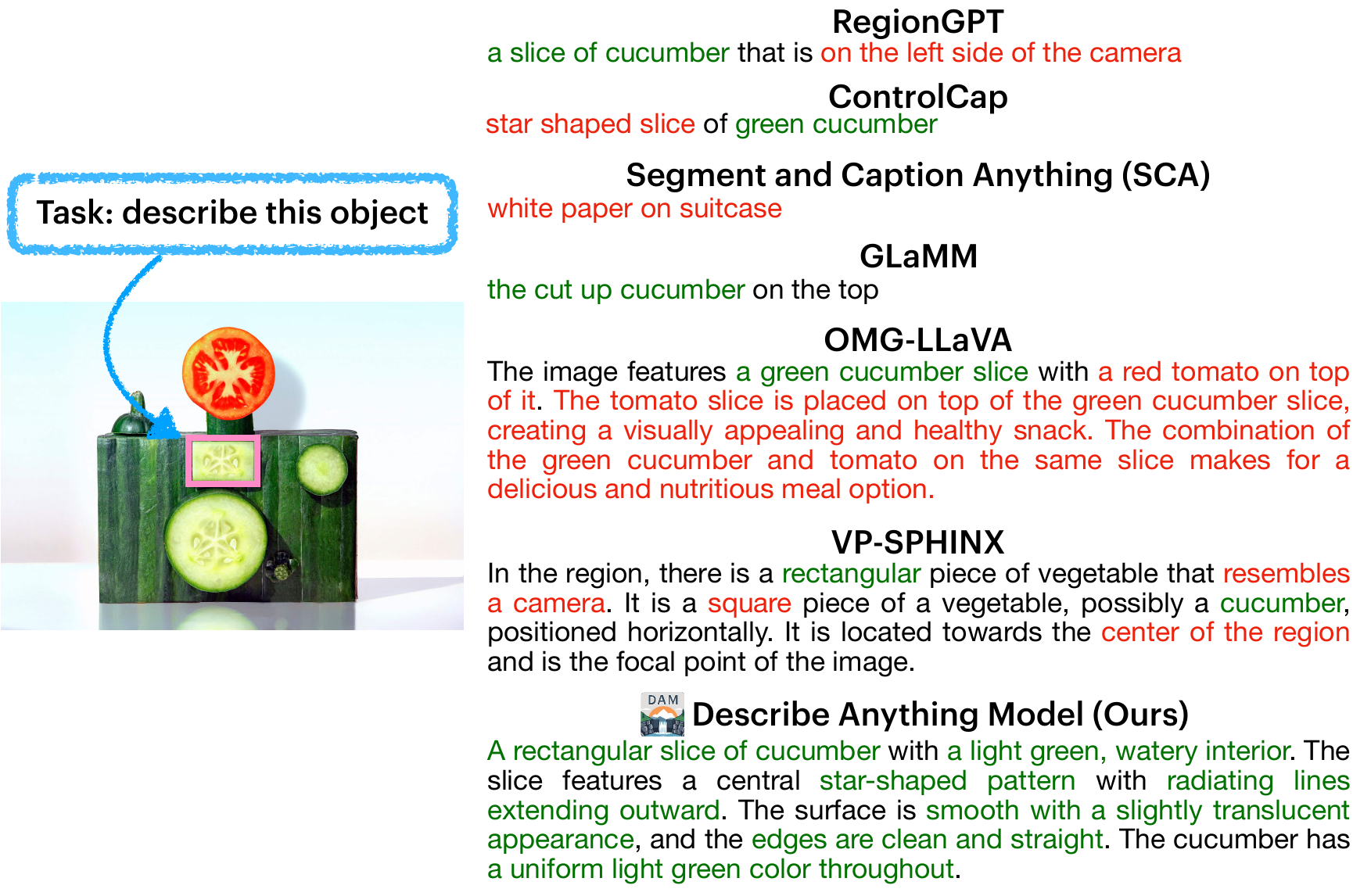}
\vspace{-6mm}
\caption{\ModelName{} generates detailed and localized descriptions, whereas prior works generate descriptions that are less precise. \textcolor[RGB]{1,113,0}{Green}: correct description. \textcolor[RGB]{238,34,12}{Red}: factual error or mislocalization.\vspace{-2mm}}
\label{fig:comparison}
\end{figure}
}

\def\figVideoExamples#1{
\begin{figure}[#1]
\centering
\begin{subfigure}{1.0\linewidth}
    \centering
    \includegraphics[width=1.0\linewidth]{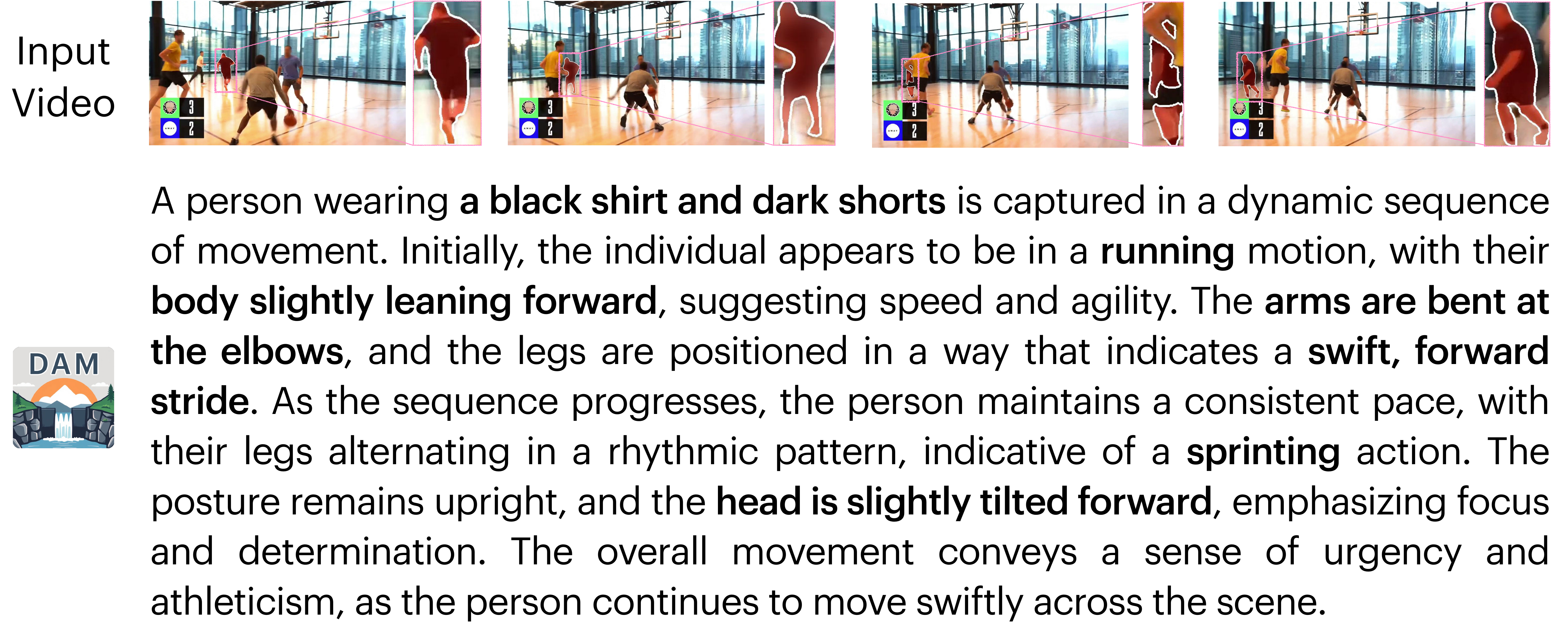}
    \caption{DAM reliably describes user-specified objects in videos, even under strong camera and object motion and heavy occlusion.}
\end{subfigure}
\vspace{6mm}
\begin{subfigure}{1.0\linewidth}
    \centering
    \includegraphics[width=1.0\linewidth]{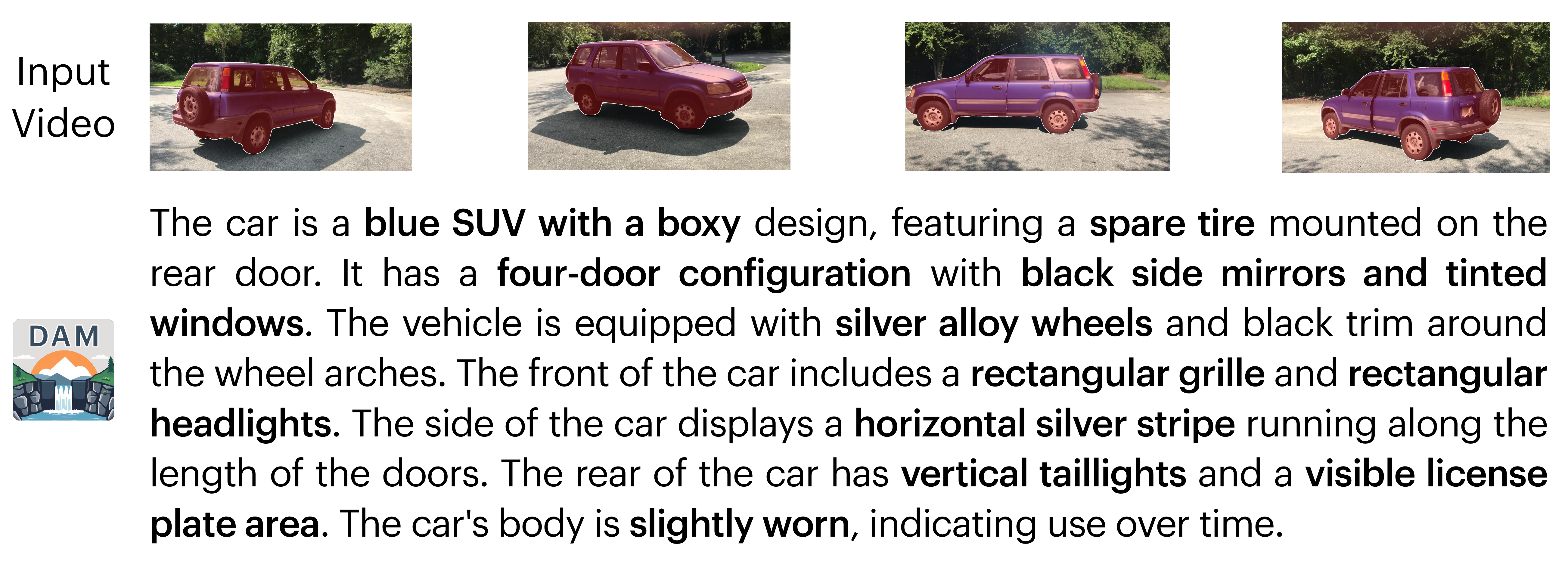}
    \caption{DAM can describe objects in multi-view datasets, like this car from Co3Dv2~\cite{reizenstein21co3d}, by integrating information from multiple frames.}
\end{subfigure}
\vspace{-12.5mm}
\caption{\textbf{\ModelName{} accurately describes user-specified regions in \textit{videos} and \textit{multi-view scenes} under challenging conditions.} More results presented in~\cref{fig:additional_video_examples}.}
\label{fig:video_examples}
\end{figure}
}

\def\figControl#1{
\begin{figure}[#1]
\centering
\vspace{-2mm}
\includegraphics[width=0.95\linewidth]{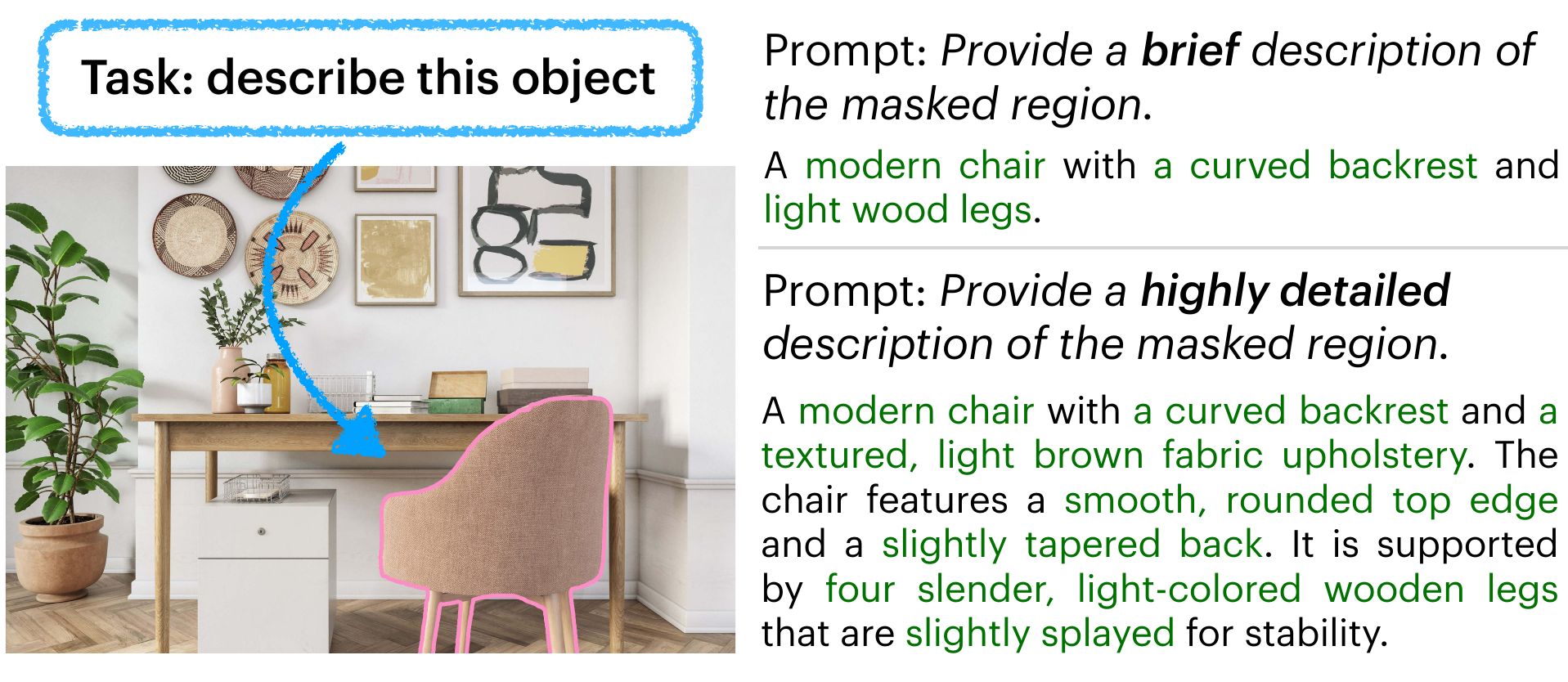}
\vspace{-4mm}
\caption{\textbf{\ModelName{} offers multi-granular localized descriptions}.}
\label{fig:control}
\end{figure}
}

\def\figQA#1{
\begin{figure}[#1]
\centering
\vspace{-2mm}
\includegraphics[width=0.95\linewidth]{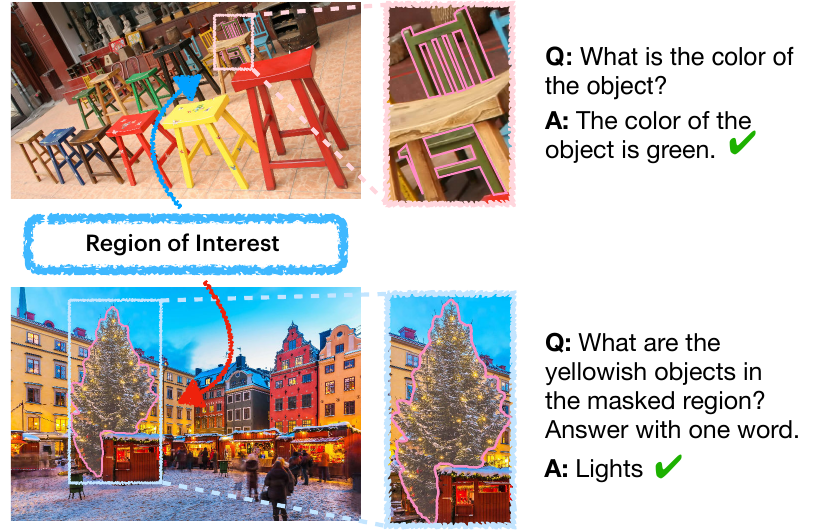}
\vspace{-2mm}
\caption{\textbf{\ModelName{} has emerging zero-shot QA capabilities} despite not fine-tuned on any regional QA datasets.}
\vspace{-2mm}
\label{fig:qa}
\end{figure}
}

\def\figLocalizedInputs#1{
\begin{figure*}[#1]
\centering
\includegraphics[width=0.9\linewidth]{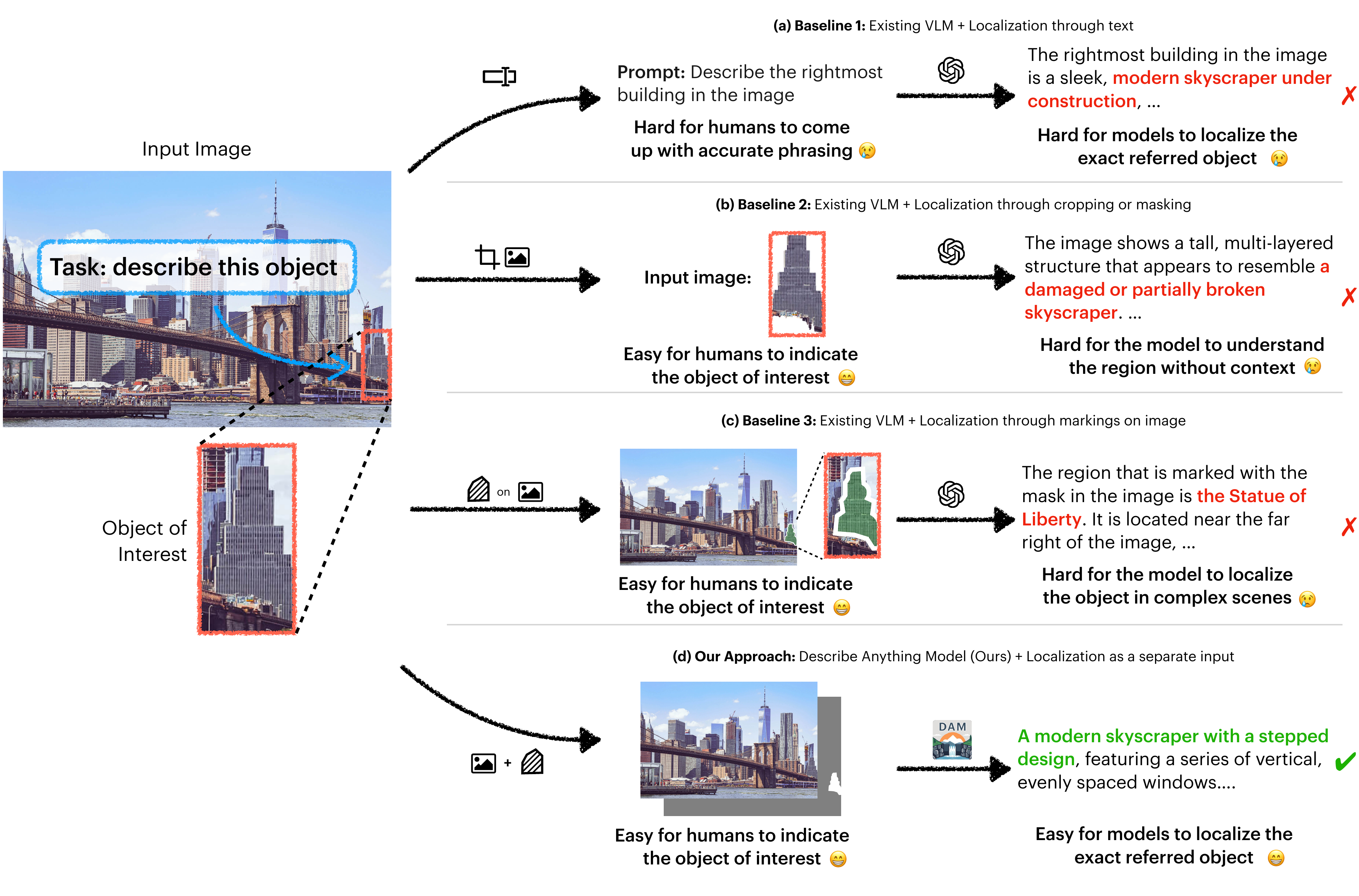}
\vspace{-2mm}
\caption{\textbf{Existing Vision-Language Models (VLMs) do not perform well in generating localized descriptions.} (a) to (c) demonstrate several ways to prompt existing VLMs, but none achieves satisfactory performance, leading to the need for a new method that is capable of providing detailed and localized descriptions. In (d), we propose a model that accepts the condition in a separate form of input, making it easy for users to specify the object of interest and for the models to accurately localize the referred object. Note that our focal prompt, proposed in \cref{sec:model}, is considered part of the Describe Anything Model and is not shown in the figure for simplicity.}
\label{fig:localized_inputs}
\end{figure*}
}

\def\figBenchmarkExample#1{
\begin{figure*}[#1]
\centering
\includegraphics[width=1.0\linewidth]{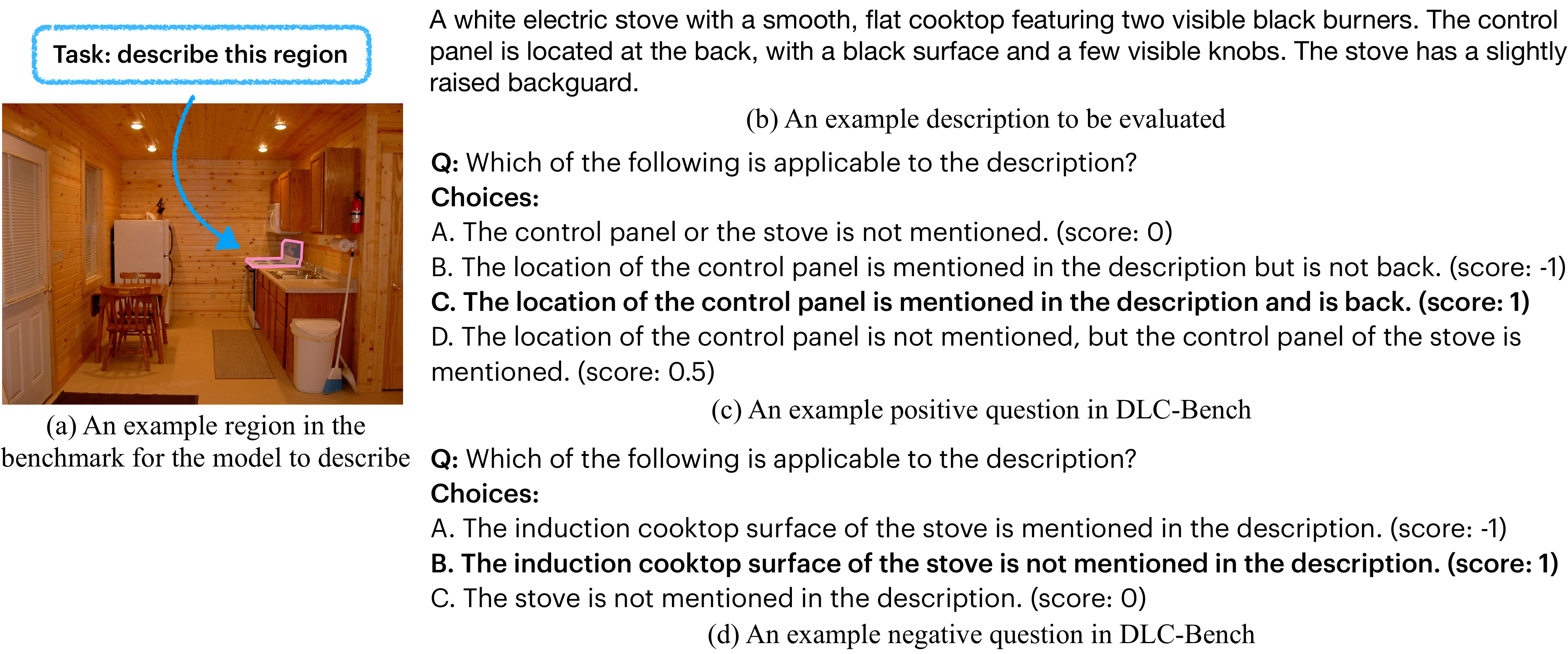}
\vspace{-2mm}
\caption{\textbf{An example from DLC-Bench for detailed localized captioning.} \textbf{(a)} The process begins by prompting a model to describe a specified region within the image. The resulting description is then evaluated using a text-only LLM as a judge that rates each response by answering positive and negative questions. \textbf{(b)} shows an example description to be evaluated.
\textbf{(c)} Positive questions are designed to test whether the model correctly identifies specific details within the described region. The model receives points for accurate details and is penalized for factual errors. The bold option (option C) indicates that the LLM judge believes that option C is applicable, allowing the model to get a point for this example positive question.
\textbf{(d)} Negative questions ensure the model refrains from mentioning irrelevant or nonexistent details. Mislocalization or hallucination results in penalties to prevent false positives. The bold option (option B) indicates that the LLM judge believes that option B is applicable, allowing the model to get a point for this negative question.}
\label{fig:benchmark_example}
\end{figure*}
}

\def\tabSOMResult#1{
\begin{table}[#1]
    \centering
    \setlength{\tabcolsep}{1.5pt}
    \begin{adjustbox}{width=1.02\linewidth,center}
    \begin{tabular}{lc@{\hspace{2.5mm}}c@{\hspace{3.0mm}}c@{\hspace{3.0mm}}c}
        \toprule
        {\small \textbf{Method}} & {\small \hspace{-5mm}\textbf{\#Params}} & {\small \textbf{Pos (\%)}} & {\small \textbf{Neg (\%)}} & {\small \textbf{Avg (\%)}} \\
        \midrule
        \multicolumn{5}{l}{\textit{API-only General VLMs:}} \\
        GPT-4o (SOM)~\cite{gpt4o} & - & 5.0 & 29.2 & 17.1 \\
        o1 (SOM)~\cite{o1}$^\dagger$ & - & 0.8 & 28.0 & 14.4 \\
        Claude 3.7 Sonnet (SOM)~\cite{team2025claude}$^\dagger$ & - & 0.5 & 40.2 & 20.4 \\
        Gemini 2.5 Pro (SOM)~\cite{team2023gemini,team2025gemini}$^\dagger$ & - & 13.2 & 65.0 & 39.1 \\
        \midrule
        \multicolumn{5}{l}{\textit{Open-source General VLMs:}} \\
        Llama-3.2 Vision (SOM)~\cite{dubey2024llama}\hspace{-5mm} & 11B & 16.8 & 40.4 & 28.6 \\
        Llama-3 VILA1.5 (SOM)~\cite{lin2024vila}\hspace{-5mm} & 8B & 0.6 & 0.6 & 0.6 \\
        InternVL2.5 (SOM)~\cite{chen2023internvl,chen2024expanding,wang2024mpo} & 8B & 8.6 & 28.6 & 18.6 \\
        LLaVA v1.6 (SOM)~\cite{liu2024visual,liu2023improvedllava,liu2024llavanext} & 7B & 2.2 & 3.8 & 3.0 \\
        Qwen2.5-VL (SOM)~\cite{wang2024qwen2,Qwen2.5-VL} & 7B & 8.5 & 27.2 & 17.8 \\
        VILA1.5 (SOM)~\cite{lin2024vila} & 3B & -0.4 & 15.4 & 7.5 \\
        \midrule
        \textbf{DAM (Ours)} & 3B & \textbf{52.3} & \textbf{82.2} & \textbf{67.3} \\
        \bottomrule
    \end{tabular}
    \end{adjustbox}
    \caption{\textbf{Additional results with existing general VLMs using Set-of-Mark (SoM) prompting~\cite{yang2023set}.} The results are accuracies on detailed localized captioning in DLC-Bench. Compared with results in~\cref{tab:main_result} which are obtained with the same prompt engineering as we use in the stage 1 of our data pipeline, SoM leads to degraded quality. In this comparison, the advantages of our method, compared with prior baselines, are much larger. Negative numbers are due to penalties from factual errors. Note that region-specific VLMs, including our proposed \ModelName{}, have predefined ways of inputting regions, and thus SoM prompting is not applicable to these models. $\dagger$: models with thinking mode.}
    \label{tab:som_result}
\end{table}
}

\def\tabDatasets#1{
\begin{table}[#1]
    \centering
    \setlength{\tabcolsep}{3.5pt}
    \begin{tabular}{lrr}
    \toprule
    \textbf{Dataset} & \textbf{\# Images} & \textbf{\# Regions} \\
    \midrule
    \multicolumn{3}{l}{\textit{Stage 1:}} \\
    LVIS~\cite{gupta2019lvis} & 90,613 & 373,551 \\
    Mapillary Vistas v2.0~\cite{neuhold2017mapillary} & 17,762 & 100,538 \\
    COCO Stuff~\cite{caesar2016coco} & 28,365 & 32,474 \\
    OpenImages v7~\cite{OpenImages2,OpenImages} & 64,874 & 96,006 \\
    PACO~\cite{ramanathan2023paco} & 24,599 & 81,325 \\
    \midrule
    \multicolumn{3}{l}{\textit{Stage 2:}} \\
    SA-1B (10\%) & 592,822 & 774,309 \\
    \midrule
    \textbf{Total} & \textbf{819,035} & \textbf{1,458,203} \\
    \bottomrule
    \end{tabular}
    \caption{\textbf{Dataset statistics across stages with total images and regions for training detailed localized image captioning.} In stage 1, we annotated 684k regions across 226k images from existing instance and semantic segmentation datasets. In stage 2, we perform SSL on 10\% of SA-1B images without using the masks provided by the datasets, resulting in 774k regions across 593k images. In total, we annotated 1.46M regions across 819k images with detailed localized descriptions. This diverse and high-quality dataset is the key to our model's performance. Note that due to filtering the number of instances and images are lower than the number of instances and images in the original dataset.}
\label{tab:datasets}
\end{table}
}

\def\tabVideoDatasets#1{
\begin{table}[#1]
    \centering
    \setlength{\tabcolsep}{3.5pt}
    \begin{tabular}{lrr}
    \toprule
    \textbf{Dataset} & \textbf{\# Videos} & \textbf{\# Regions} \\
    \midrule
    SA-V~\cite{ravi2024sam} & 36,922 & 93,969 \\
    \bottomrule
    \end{tabular}
    \caption{\textbf{Dataset statistics across stages with total videos and regions for training detailed localized \textit{video} captioning.} We label 94k regions across 37k videos from SA-V dataset~\cite{ravi2024sam} for detailed localized video captioning. Note that each region indicates an instance across multiple frames in the video.}
\label{tab:video_datasets}
\end{table}
}

\def\figAdditionalQA#1{
\begin{figure}[#1]
\centering
\vspace{-2mm}
\includegraphics[width=1.0\linewidth]{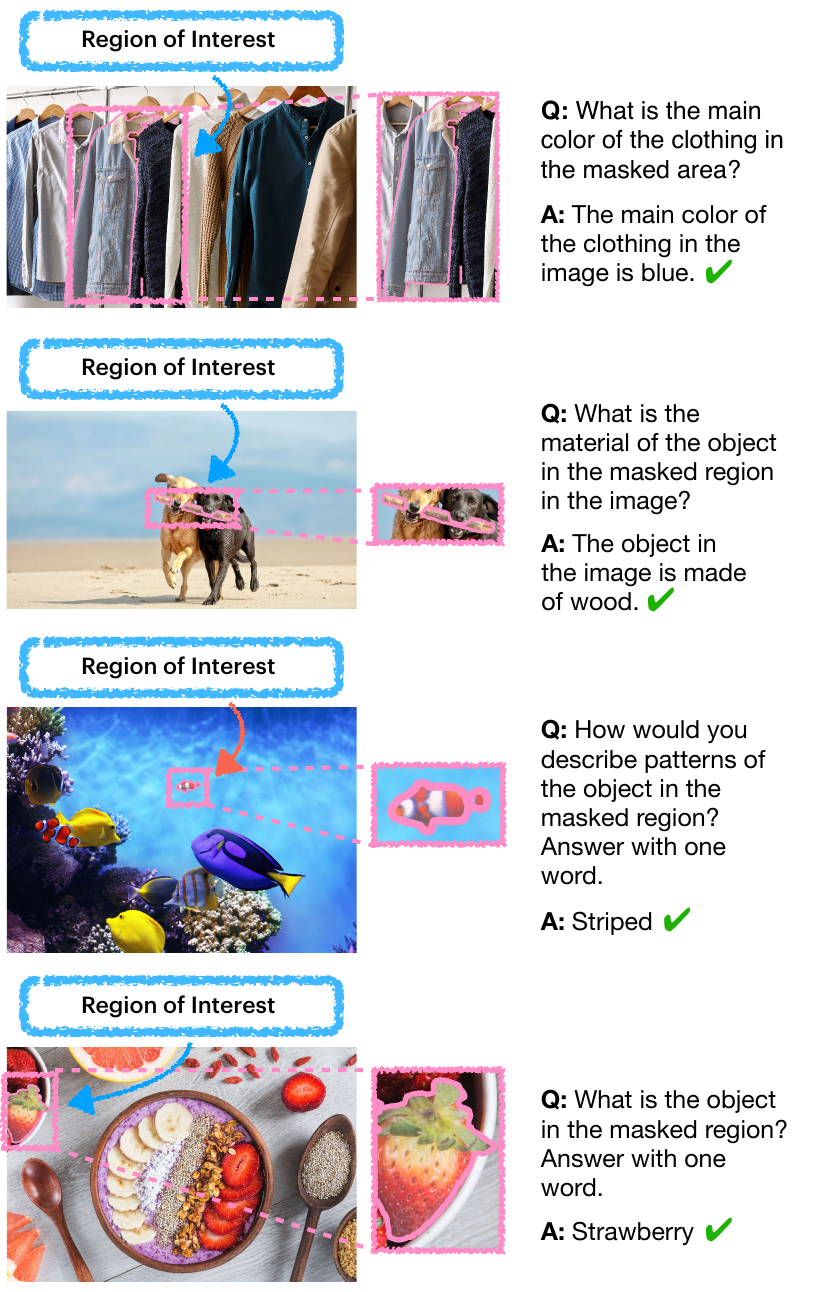}
\vspace{-2mm}
\caption{\textbf{Emerging zero-shot QA capabilities}. \ModelName{} could answer questions about regions in an image, showcasing capabilities such as object recognition and property identification.}
\label{fig:qa2}
\end{figure}
}

\def\tabAblationPromptAug#1{
\begin{table*}[#1]
    \centering
    \setlength{\tabcolsep}{3.5pt}
    \begin{tabular}{lccc}
        \toprule
        \textbf{Prompt Augmentation} & \textbf{Pos (\%)} & \textbf{Neg (\%)} & \textbf{Avg (\%)} \\
        \midrule
        No & \textbf{52.3} & \textbf{82.2} & \textbf{67.3} \\
        Yes & 51.3 & \textbf{82.2} & 66.7 \\
        \bottomrule
    \end{tabular}
    \caption{\textbf{Comparison of performance of \ModelName{} with and without prompt augmentation.} Prompt augmentation has minimal effect on \ModelName{}'s performance on DLC-Bench. While descriptions generated by the model may occasionally be less detailed, leading to a slight decrease in the performance on positive questions, we observed that prompt augmentation enhances instruction following when prompts include specific guidelines, such as length constraints. We use the model without prompt augmentation with our benchmark, including ablations, by default.}
\label{tab:ablation_prompt_aug}
\end{table*}
}

\def\tabAblationJointTraining#1{
\begin{table*}[#1]
    \centering
    \setlength{\tabcolsep}{3.5pt}
    \begin{tabular}{lccc}
        \toprule
        \textbf{Setting} & \textbf{Pos (\%)} & \textbf{Neg (\%)} & \textbf{Avg (\%)} \\
        \midrule
        Image-only Training & 52.3 & 82.2 & 67.3 \\
        Image+Video Joint Training & \textbf{52.4} & \textbf{85.4} & \textbf{68.9} \\
        \bottomrule
    \end{tabular}
    \caption{\textbf{Comparison of performance of our image-only \ModelName{} and \ModelName{} trained with both localized image description task and localized video description task.} Joint training benefits generating high-quality localized image descriptions compared to image-only training.}
\label{tab:ablation_joint_training}
\end{table*}
}

\def\figAdditionalImageExamples#1{
\begin{figure}[#1]
\centering
\vspace{-2mm}
\includegraphics[width=1.0\linewidth]{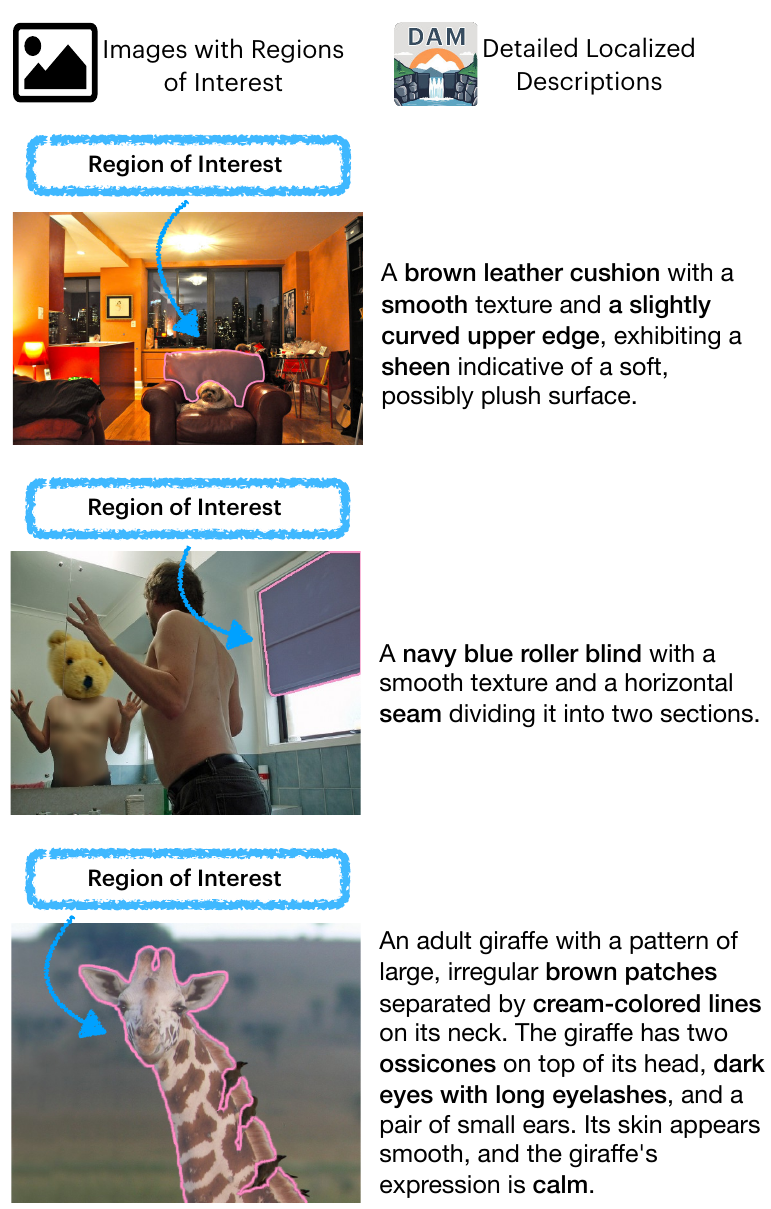}
\vspace{-2mm}
\caption{\textbf{Additional results from LVIS~\cite{gupta2019lvis} demonstrating \ModelName{}'s detailed localized image captioning capabilities}. Our model exhibits robust region understanding and localization across diverse scenarios. It produces precise descriptions of objects within masked regions while successfully identifying challenging details like the roller blind in the second example through effective use of contextual cues.}
\label{fig:additional_image_examples}
\end{figure}
}

\def\figComparisonWithGPTVideo#1{
\begin{figure*}[#1]
\centering
\vspace{-2mm}
\includegraphics[width=0.95\linewidth]{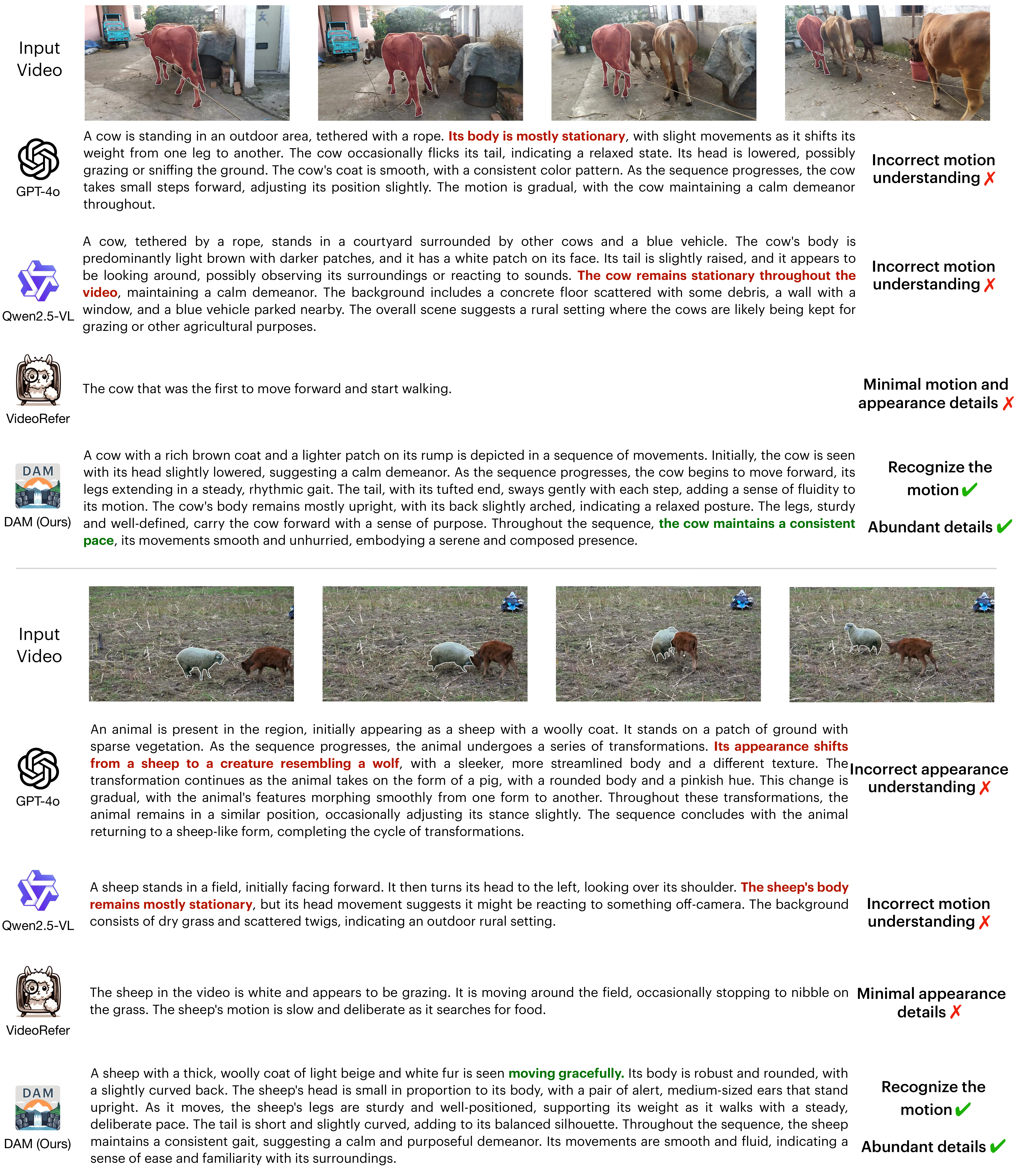}
\vspace{-2mm}
\caption{\textbf{Our proposed \ModelName{} demonstrates superior localized video understanding compared to GPT-4o~\cite{gpt4o}, QwenVL-2.5~\cite{Qwen2.5-VL}, and VideoRefer~\cite{yuan2024videorefer}}. \textbf{Top figure:} DAM accurately captures the cow's forward movement with comprehensive details, whereas GPT-4o and QwenVL-2.5 mistakenly perceive the cow as stationary. Compared to VideoRefer, DAM provides richer descriptions of both motion and appearance. \textbf{Bottom figure:} DAM correctly recognizes the animal as a sheep and accurately describes its graceful movement, while GPT-4o erroneously identifies it as transforming into other animals, and QwenVL-2.5 incorrectly perceives that only the sheep's head is moving. VideoRefer provides limited appearance details, while DAM offers extensive, accurate descriptions. These cases highlight DAM's precise understanding of motion and appearance throughout video sequences.}
\label{fig:comparison_with_gpt4o_video}
\end{figure*}
}

\def\figComparisonWithGPTImage#1{
\begin{figure*}[#1]
\centering
\vspace{-2mm}
\includegraphics[width=1.0\linewidth]{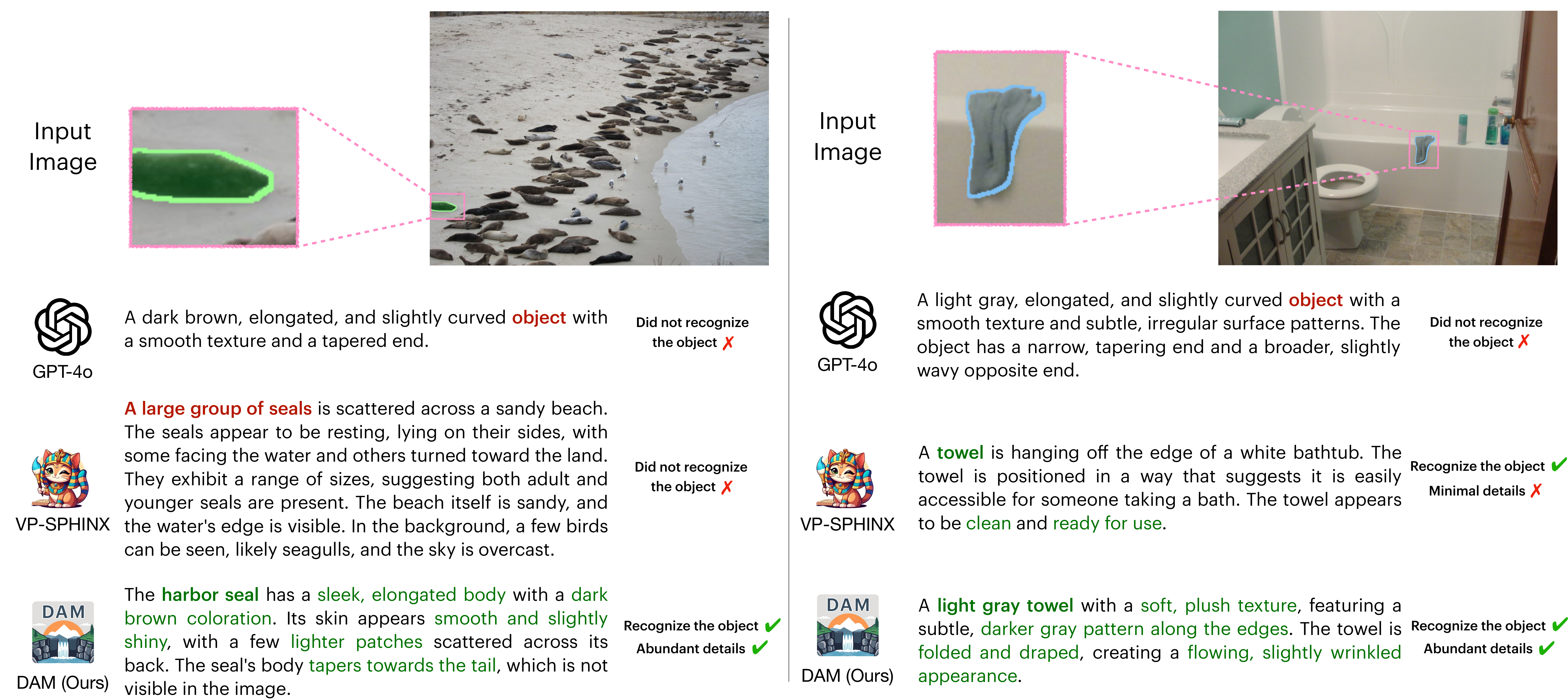}
\vspace{-2mm}
\caption{\textbf{Qualitative comparisons demonstrate the superior localized image understanding capabilities of our model compared to GPT-4o~\cite{gpt4o} and VP-SPHINX~\cite{lin2024draw}, our strongest open-weight baseline}. GPT-4o struggles to recognize objects in masked regions accurately, offering only vague descriptions. In the left image, VP-SPHINX incorrectly describes a group of seals when the masked region contains only one seal. In the right image, VP-SPHINX identifies the towel but provides minimal detail, missing key attributes like its color. In contrast, our model delivers precise, detailed descriptions and captures the seal's sleek elongated body, dark brown coloration with lighter patches, and the towel's light gray color, wrinkled texture, and darker edge pattern. This superior performance stems from our model's architecture that effectively fuses object-specific details with broader contextual understanding.}
\label{fig:comparison_with_gpt4o_image}
\end{figure*}
}

\def\figAdditionalVideoExamples#1{
\begin{figure*}[#1]
\centering
\vspace{-2mm}
\includegraphics[width=1.0\linewidth]{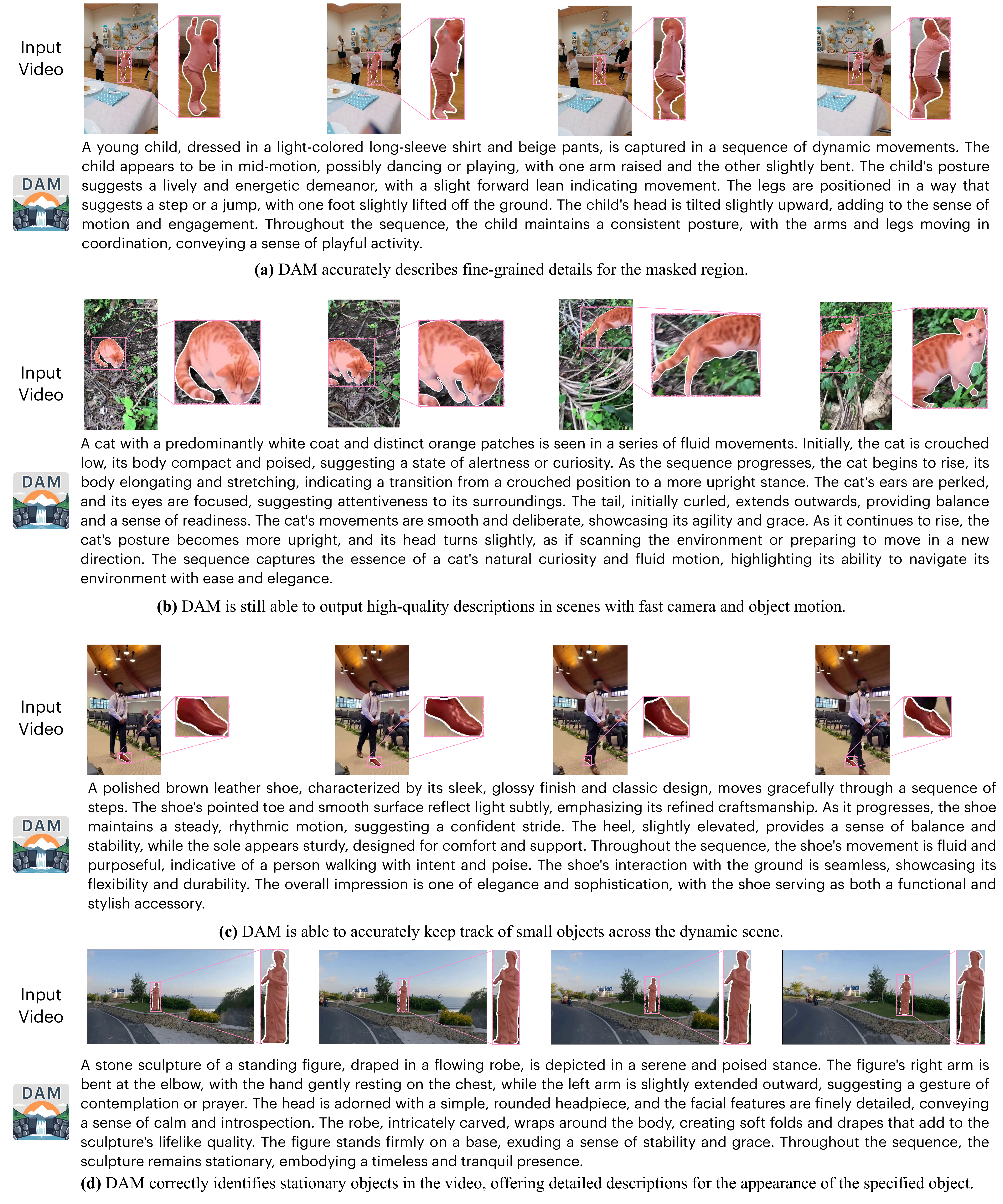}
\vspace{-2mm}
\caption{\textbf{Additional results from \ModelName{} on detailed localized video captioning (Part 1)}. Our model is able to accurately describe small objects in complex scenes that involve large object motion and camera motion. Our model also correctly identifies stationary objects captured by a non-stationary camera by saying they are stationary. Videos visualized in this figure are from SA-V~\cite{ravi2024sam} dataset.}
\label{fig:additional_video_examples}
\end{figure*}
}

\def\figAdditionalVideoExamplesPartTwo#1{
\begin{figure*}[#1]
\centering
\vspace{-2mm}
\includegraphics[width=1.0\linewidth]{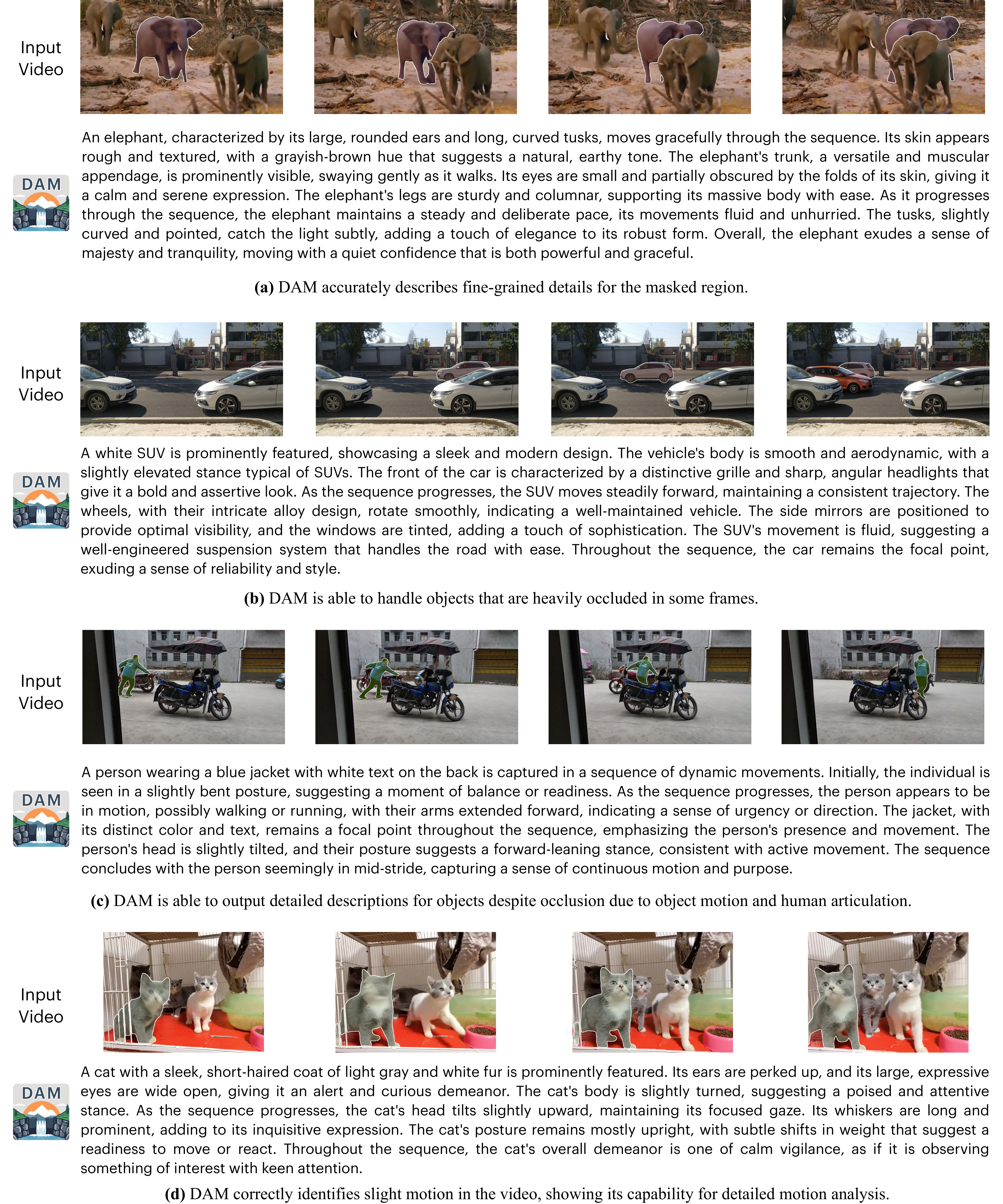}
\vspace{-2mm}
\caption{\textbf{Additional results from \ModelName{} on detailed localized video captioning (Part 2)}. Our model is able to accurately describe objects that are partially occluded and is able to perceive and describe slight motion. Videos visualized in this figure are from MOSE~\cite{ding2023mose} dataset.}
\label{fig:additional_video_examples_part2}
\end{figure*}
}

\def\figFailureCase#1{
\begin{figure*}[#1]
\centering
\begin{subfigure}{0.95\linewidth}
    \centering
    \includegraphics[width=\linewidth]{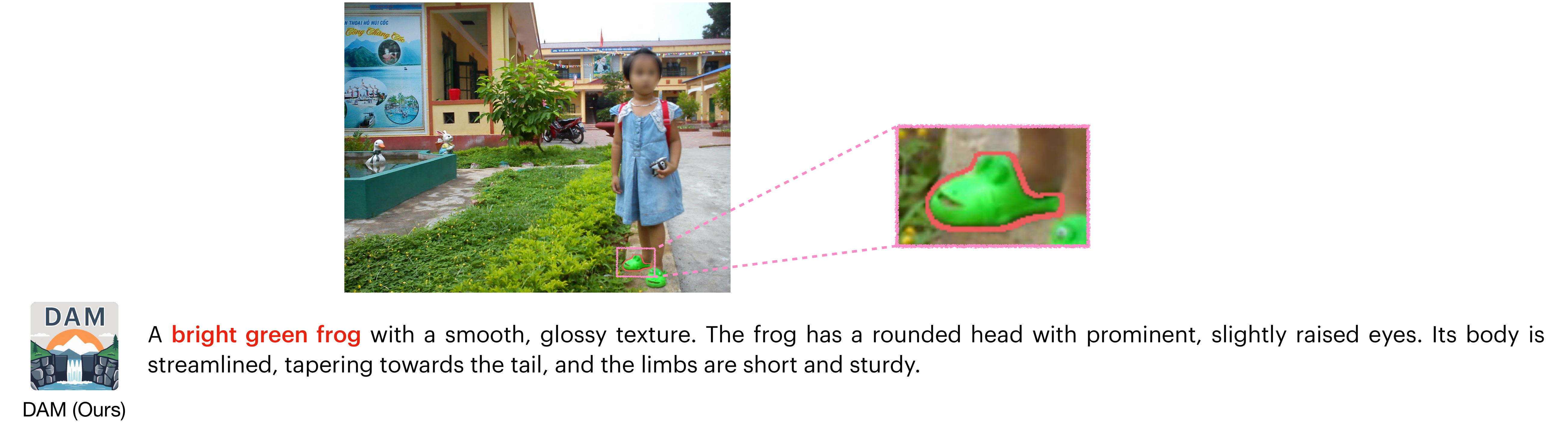}
    \caption{DAM might still misrecognize a region and output an incorrect description. For example, it misrecognizes the frog-shaped slipper to be a frog.}
    \label{fig:failure_case_a}
\end{subfigure}
\vspace{4mm}
\begin{subfigure}{0.95\linewidth}
    \centering
    \includegraphics[width=\linewidth]{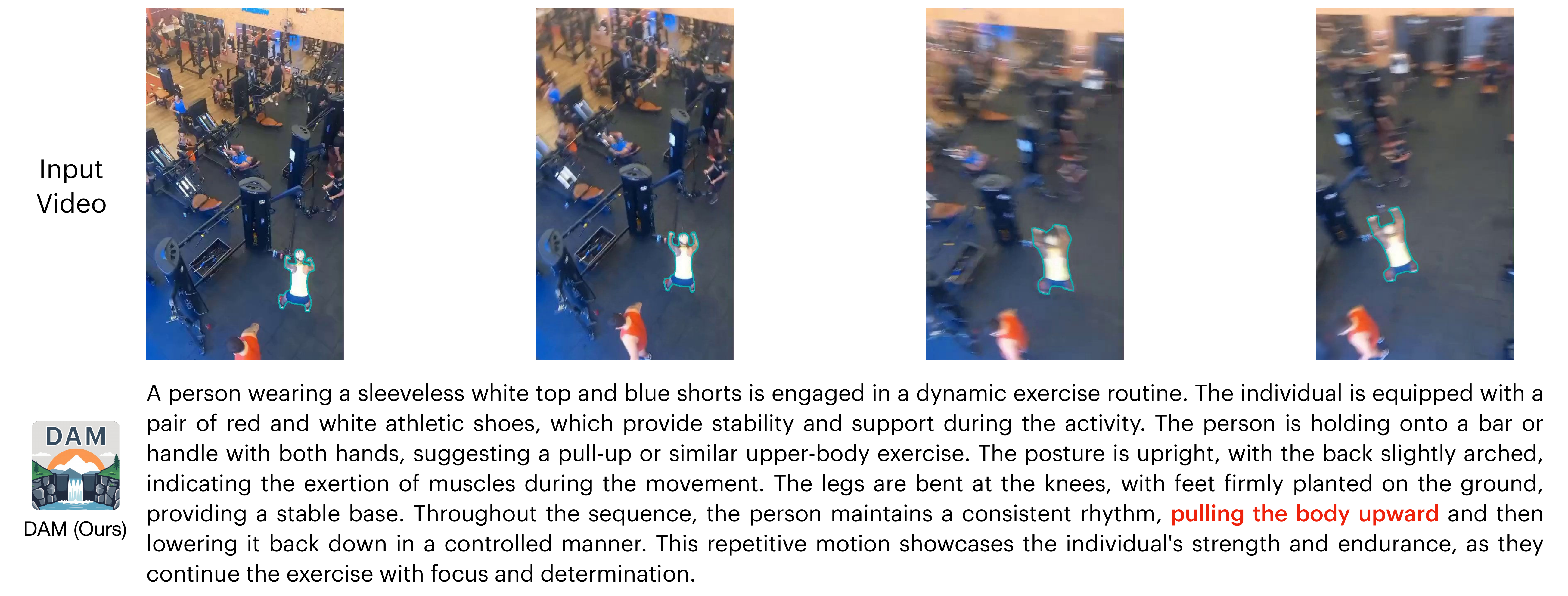}
    \caption{DAM might also be confused by the combination of the object motion and the camera motion. In this example, it makes the mistake of describing the person as pulling the body upward.}
    \label{fig:failure_case_b}
\end{subfigure}
\vspace{-2mm}
\caption{Failure cases for our proposed \ModelName{}.}
\label{fig:failure_case}
\end{figure*}
}

\def\tabSemMetrics#1{
    \centering
    \setlength{\tabcolsep}{4pt}
    \begin{adjustbox}{width=1.0\linewidth,center}
    \begin{tabular}{lcccc}
        \toprule
        {\small \textbf{Method}} & \multicolumn{2}{c}{\small \textbf{LVIS (\%)}} & \multicolumn{2}{c}{\small \textbf{PACO (\%)}} \\
        \cmidrule(lr){2-3} \cmidrule(lr){4-5}
        & {\small \textbf{Sem. Sim.} ($\uparrow$)} & {\small \textbf{Sem. IoU} ($\uparrow$)} & {\small \textbf{Sem. Sim.} ($\uparrow$)} & {\small \textbf{Sem. IoU} ($\uparrow$)} \\
        \midrule
        LLaVA-7B~\cite{liu2024visual} & 49.0 & 19.8 & 42.2 & 14.6 \\
        Shikra-7B~\cite{chen2023shikra} & 49.7 & 19.8 & 43.6 & 11.4 \\
        GPT4RoI-7B~\cite{zhang2023gpt4roi} & 51.3 & 12.0 & 48.0 & 12.1 \\
        Osprey-7B~\cite{yuan2024osprey} & 65.2 & 38.2 & 73.1 & \underline{52.7} \\
        Ferret-13B~\cite{you2023ferret} & 65.0 & 37.8 & - & - \\
        VP-SPHINX-7B~\cite{lin2024draw} & 86.0 & 61.2 & 74.2 & 49.9 \\
        VP-LLAVA-8B~\cite{lin2024draw} & \underline{86.7} & \underline{61.5} & \underline{75.7} & 50.0 \\
        \textbf{\ModelName{}-8B (Ours)} & \textbf{89.0} & \textbf{77.7} & \textbf{84.2} & \textbf{73.2} \\
        \bottomrule
    \end{tabular}
    \end{adjustbox}
    \vspace{-2mm}
    \caption{LVIS~\cite{gupta2019lvis} and PACO~\cite{ramanathan2023paco} open-class \textbf{keyword-level} captioning benchmarks. \ModelName{} excels particularly in the challenging PACO benchmark that requires distinguishing between objects and parts.}
    \label{tab:keyword_lvis_paco}
}

\def\tabFlickr#1{
\centering
    \setlength{\tabcolsep}{4pt}
    \vspace{1mm}
    \begin{adjustbox}{width=1.0\linewidth,center}
    \begin{tabular}{lccccc}
        \toprule
        \textbf{Method} & \textbf{BLEU} & \textbf{METEOR} & \textbf{ROUGE-L} & \textbf{CIDEr} & \textbf{SPICE} \\
        \midrule
        Shikra-7B~\cite{chen2023shikra} & 18.2 & 15.3 & 25.2 & 49.8 & 22.0 \\
        GPT4RoI-7B~\cite{zhang2023gpt4roi} & 19.7 & 17.7 & 29.9 & 61.7 & 24.0 \\
        Ferret-7B~\cite{you2023ferret} & 11.1 & 8.8 & 22.7 & 38.1 & 17.5 \\
        VP-SPHINX-13B~\cite{lin2024draw} & 15.2 & 15.6 & 27.2 & 67.4 & 24.0 \\
        RegionGPT-7B~\cite{guo2024regiongpt} & 16.1 & 16.7 & 27.4 & 54.6 & 20.5 \\
        \textbf{\ModelName{}-8B (Ours)} & \textbf{22.6} & \textbf{17.8} & \textbf{31.2} & \textbf{74.7} & \textbf{25.5} \\
        \bottomrule
    \end{tabular}
    \end{adjustbox}
    \vspace{-2mm}
    \caption{Zero-shot evaluation on \textbf{phrase-level} dataset Flickr30k Entities~\cite{plummer2015flickr30k}. Our model achieves 12.3\% average relative improvement against previous best.}
    \label{tab:flickr30k}
}

\def\tabRef#1{
    \centering
    \setlength{\tabcolsep}{4pt}
    \vspace{-5.5mm}
    \begin{adjustbox}{width=1.0\linewidth,center}
    \begin{tabular}{lcccccc}
        \toprule
        & \multicolumn{5}{c}{\small \textbf{Short Captioning Metrics}} & \multicolumn{1}{c}{\small \textbf{Long Cap. Metrics}} \\
        \cmidrule(lr){2-6} \cmidrule(lr){7-7}
        {\small \textbf{Method}} & {\small \textbf{BLEU}} & {\small \textbf{METEOR}} & {\small \textbf{ROUGE-L}} & {\small \textbf{CIDEr}} & {\small \textbf{SPICE}} & {\small \textbf{CLAIR}} \\
        \midrule
        Shikra-7B~\cite{chen2023shikra} & 29.5 & 11.1 & 23.9 & 42.7 & 9.0 & 34.5 \\
        GPT4RoI-7B~\cite{zhang2023gpt4roi} & 27.1 & 11.6 & 26.8 & 59.9 & 11.1 & 43.9 \\
        Ferret-7B~\cite{you2023ferret} & 24.6 & 10.7 & 22.3 & 39.7 & 8.2 & 45.2 \\
        GLaMM-7B~\cite{rasheed2024glamm} & 23.2 & 10.1 & 23.8 & 51.1 & 8.7 & 43.8 \\
        VP-SPHINX-13B~\cite{lin2024draw} & 22.6 & 10.7 & 22.6 & 32.4 & 7.6 & 51.2 \\
        RegionGPT-7B~\cite{guo2024regiongpt} & 25.4 & 12.2 & 25.3 & 42.0 & 8.1 & 37.2 \\
        \textbf{\ModelName{}-8B (Ours)} & \textbf{38.7} & \textbf{19.4} & \textbf{37.1} & \textbf{70.0} & \textbf{16.9} & \textbf{57.9} \\
        \bottomrule
    \end{tabular}
    \end{adjustbox}
    \vspace{-2mm}
    \caption{Zero-shot evaluation on the \textbf{detailed captioning} dataset Ref-L4~\cite{chen2024revisiting}. Our method achieves 33.4\% and 13.1\% average relative improvement on the short/long language-based captioning metrics, respectively.\vspace{-6mm}}
    \label{tab:ref_l4}
}

\def\tabVideoRefer#1{
\begin{table}[#1]
    \centering
    \setlength{\tabcolsep}{4pt}
    \begin{adjustbox}{width=0.7\linewidth,center}
    \begin{tabular}{lccccc}
        \toprule
        {\small \textbf{Method}} & {\small \textbf{SC}} & {\small \textbf{AD}} & {\small \textbf{TD}} & {\small \textbf{HD}$\dagger$} & {\small \textbf{Avg.}} \\
        \midrule
        \multicolumn{5}{l}{\textit{Zero-shot:}} \\
        Qwen2-VL-7B~\cite{wang2024qwen2} & 3.30 & 2.54 & 2.22 & 2.12 & 2.55 \\
        InternVL2-26B~\cite{chen2023internvl} & 4.08 & \textbf{3.35} & 3.08 & 2.28 & 3.20 \\
        GPT-4o-mini~\cite{gpt4o} & 3.89 & 3.18 & 2.62 & 2.50 & 3.05 \\
        GPT-4o~\cite{gpt4o} & 4.15 & 3.31 & \textbf{3.11} & 2.43 & 3.25 \\
        Osprey-7B~\cite{yuan2024osprey} & 3.30 & 2.66 & 2.10 & 1.58 & 2.41 \\
        Ferret-7B~\cite{you2023ferret} & 3.20 & 2.38 & 1.97 & 1.38 & 2.23 \\
        Elysium-7B~\cite{wang2024elysium} & 2.35 & 0.30 & 0.02 & \textbf{3.59} & 1.57 \\
        Artemis-7B~\cite{qiu2025artemis} & 3.42 & 1.34 & 1.39 & 2.90 & 2.26 \\
        \textbf{\ModelName{}-8B (Ours)} & \textbf{4.45} & 3.30 & 3.03 & 2.58 & \textbf{3.34} \\
        \midrule
        \multicolumn{5}{l}{\textit{In-domain*:}} \\
        VideoRefer-7B~\cite{yuan2024videorefer} & 4.44 & 3.27 & 3.10 & 3.04 & 3.46 \\
        \textbf{\ModelName{}-8B (Ours)} & \textbf{4.69} & \textbf{3.61} & \textbf{3.34} & \textbf{3.09} & \textbf{3.68} \\
        \bottomrule
    \end{tabular}
    \end{adjustbox}
    \vspace{-2mm}
    \caption{\textbf{Performance on detailed localized video description on VideoRefer-Bench-D~\cite{yuan2024videorefer}.} $\dagger$: We provide analysis on hallucination scores (HD) in~\cref{sec:quantitative_results,ssec:reference_captions}. *: trained on in-domain VideoRefer-700k with regard to VideoRefer-Bench, both sourcing videos from Panda-70M~\cite{chen2024panda}.}
    \label{tab:videorefer}
\end{table}
}

\def\tabHCSTVG#1{
\begin{table}[#1]
    \centering
    \setlength{\tabcolsep}{4pt}
    \vspace{-2mm}
    \begin{adjustbox}{width=0.85\linewidth,center}
    \begin{tabular}{lccccc}
        \toprule
        {\small \textbf{Method}} & {\small \textbf{BLEU@4}} & {\small \textbf{METEOR}} & {\small \textbf{ROUGE-L}} & {\small \textbf{CIDEr}} & {\small \textbf{SPICE}} \\
        \midrule
        Osprey-7B~\cite{yuan2024osprey} & 0.7 & 12.0 & 18.0 & 1.2 & 15.6 \\
        Ferret-13B~\cite{you2023ferret} & 0.5 & 10.2 & 17.0 & 1.2 & 11.2 \\
        Shikra-7B~\cite{chen2023shikra} & 1.3 & 11.5 & 19.3 & 3.1 & 13.6 \\
        Merlin-7B~\cite{yu2024merlin} & 3.3 & 11.3 & 26.0 & 10.5 & 20.1 \\
        Artemis-7B~\cite{qiu2025artemis} & 15.5 & 18.0 & 40.8 & 53.2 & 25.4 \\
        VideoRefer-7B~\cite{yuan2024videorefer} & 16.5 & 18.7 & 42.4 & 68.6 & 28.3 \\
        \textbf{\ModelName{}-8B (Ours)} & \textbf{19.8} & \textbf{21.0} & \textbf{45.9} & \textbf{91.0} & \textbf{31.4} \\
        \bottomrule
    \end{tabular}
    \end{adjustbox}
    \vspace{-3mm}
    \caption{\textbf{Detailed localized \textit{video} captioning} on HC-STVG~\cite{tang2021human}.\vspace{-4mm}}
    \label{tab:hc_stvg}
\end{table}
}

\def\tabAblationModel#1{
\begin{table*}[#1]
    \centering
    \setlength{\tabcolsep}{4pt}
    \begin{tabular}{lcc}
    \toprule
    & VP-SPHINX Arch & Our Arch \\
    \midrule
    Avg ($\%$) & 50.2 & \textbf{63.8} \\
    \bottomrule
    \end{tabular}
    \caption{\textbf{Ablations on architecture design compared to our strongest baseline VP-SPHINX~\cite{lin2024draw}.} We trained a model with VP-SPHINX~\cite{lin2024draw} architecture on our curated DLC data from various segmentation datasets. The results on DLC-Bench indicate the advantages of our model architecture that allows detailed localized features to be presented to the LLM for DLC.}
    \label{tab:ablation_model}
\end{table*}
}

\def\tabAdvantages#1{
\begin{table*}[t]
\centering
\renewcommand{\arraystretch}{1.2}
\vspace{-4mm}
\begin{adjustbox}{width=\linewidth,center}
\begin{tabular}{p{0.12\linewidth} p{0.28\linewidth} p{0.25\linewidth} p{0.28\linewidth} p{0.25\linewidth}}
\toprule
\textbf{Component} & \textbf{Previous Practice} & \textbf{Problem} & \textbf{Our Solution} & \textbf{Advantages} \\ 
\midrule
\textbf{Describe Anything Model (\ModelName{})} & Extracting regional features from global image features & Regional details already lost in image feature extraction and not provided to the LLM & Providing \textbf{focal prompt} to proposed \textbf{localized vision backbone} & Detail-rich contextful features allowing for accurate, multi-granular localized descriptions \\
\midrule
\multirow{2}{\linewidth}{\textbf{SSL Data Pipeline (DLC-SDP)}} 
& Query a data curation VLM with referring \textbf{boxes} and \textbf{global image captions} 
& Imprecise referring to data curation model 
& Reframe the query into a \textbf{mask}-referred \textbf{keyword} expansion question
& Leverage high-quality precise human annotated regional masks and keywords \\ 
\cmidrule(lr){2-5}
& \textbf{Fully supervised learning} 
& Limited data with high annotation quality
& \textbf{Semi-supervised learning} 
& Scalable to diverse web-scale unlabeled datasets \\ 
\midrule
\textbf{Benchmark (DLC-Bench)} 
& Pred caption + \textbf{reference GT caption} $\rightarrow$ language-based similarity metrics or LLM scorer
& Incorrect hallucination penalty for correct details not present in the reference caption 
& Pred caption + \textbf{query for positive/negative attributes} $\rightarrow$ LLM scorer
& Accurate detail and hallucination assessment without relying on reference captions \\ 
\bottomrule
\end{tabular}
\end{adjustbox}
\vspace{-2mm}
\caption{Advantages of our proposed model DAM, our SSL data pipeline DLC-SDP, and our benchmark DLC-Bench to previous practices.\vspace{-2mm}}
\label{tab:advantages}
\end{table*}
}

\def\figMaskRoI#1{
\begin{figure}[#1]
\centering
\includegraphics[width=1.0\linewidth]{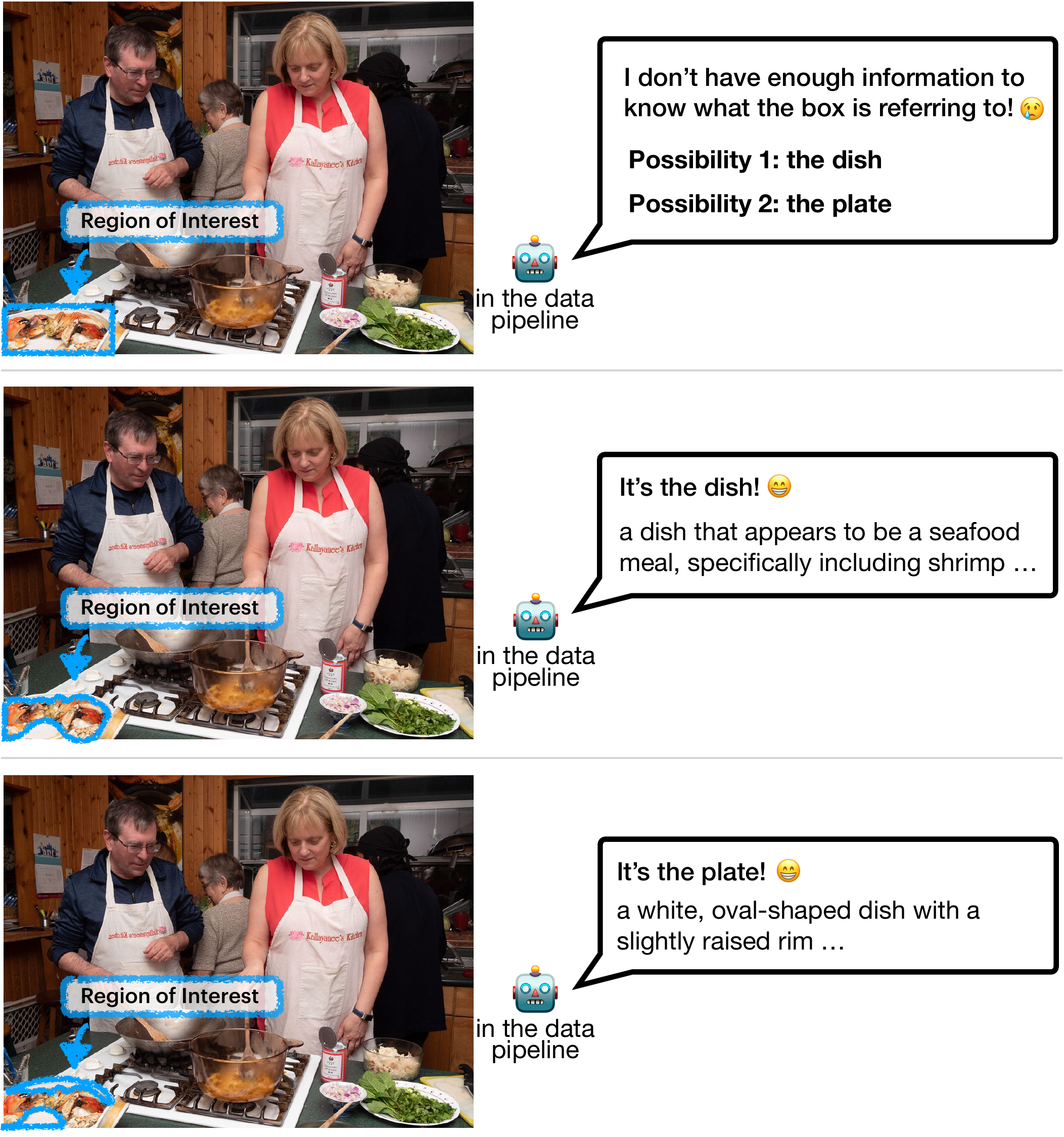}
\caption{\textbf{Caveats for using boxes to indicate region of interests.} Top: Using a box to indicate the region of interest leads to ambiguity. Middle and Bottom: Switching to a mask representation leads to more specific referring and correct descriptions.}
\label{fig:mask_roi}
\end{figure}
}

\def\figHallucinationExamples#1{
\begin{figure*}[#1]
\centering
\includegraphics[width=1.0\linewidth]{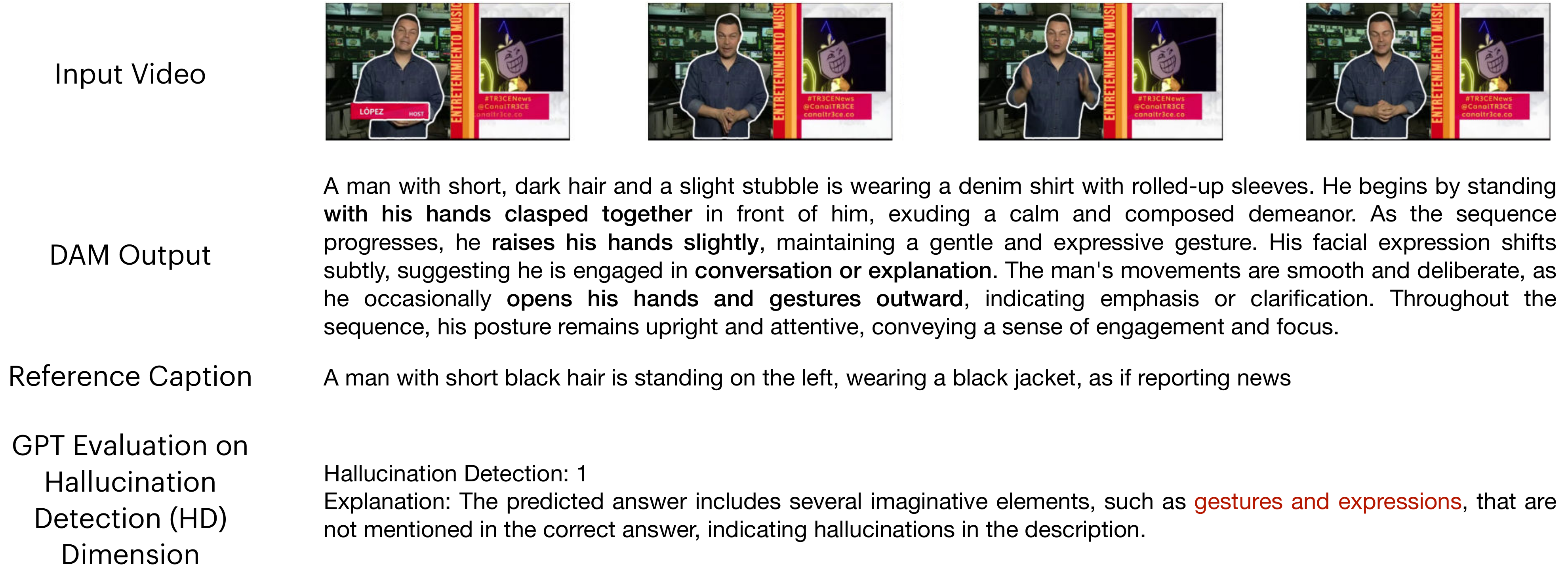}
\caption{\textbf{The pitfall of using reference captions for caption evaluation.} Evaluation benchmarks based on reference captions may incorrectly treat correct details in the predicted caption as hallucination. Since the GPT evaluator relies solely on the ground truth caption without viewing the video, it mistakenly flags gestures and expressions as hallucinations, resulting in a low score. However, the evaluation is invalid since the predicted details are correct.}
\label{fig:hallucination_examples}
\end{figure*}
}

\definecolor{codegreen}{rgb}{0,0.6,0}
\definecolor{codegray}{rgb}{0.5,0.5,0.5}
\definecolor{codepurple}{rgb}{0.58,0,0.82}
\definecolor{backcolour}{rgb}{0.95,0.95,0.92}

\lstdefinestyle{mystyle}{
    backgroundcolor=\color{backcolour},   
    commentstyle=\color{codegreen},
    keywordstyle=\color{magenta},
    numberstyle=\tiny\color{codegray},
    stringstyle=\color{codepurple},
    basicstyle=\ttfamily\footnotesize,
    breakatwhitespace=false,         
    breaklines=true,                 
    captionpos=b,                    
    keepspaces=true,                 
    numbers=left,                    
    numbersep=5pt,                  
    showspaces=false,                
    showstringspaces=false,
    showtabs=false,                  
    tabsize=2
}

\lstset{style=mystyle}

\newfloat{lstfloat}{htbp}{lop}
\floatname{lstfloat}{Algorithm}
\crefalias{lstfloat}{listing}
\def\lstfloatautorefname{Algorithm}

\lstdefinestyle{myverbatim}{
    basicstyle=\ttfamily\footnotesize,
    backgroundcolor=\color{white},
    breaklines=true,
    breakatwhitespace=true
}

\figDemo{}

\begin{abstract}
Generating detailed and accurate descriptions for specific regions in images and videos remains a fundamental challenge for vision-language models. We introduce the \textbf{Describe Anything Model (DAM)}, a model designed for detailed localized captioning (DLC). DAM preserves both local details and global context through two key innovations: a focal prompt, which ensures high-resolution encoding of targeted regions, and a localized vision backbone, which integrates precise localization with its broader context. To tackle the scarcity of high-quality DLC data, we propose a \textbf{Semi-supervised learning (SSL)-based Data Pipeline (DLC-SDP)}. DLC-SDP starts with existing segmentation datasets and expands to unlabeled web images using SSL. We introduce \textbf{DLC-Bench}, a benchmark designed to evaluate DLC without relying on reference captions. DAM sets new state-of-the-art on 7 benchmarks spanning keyword-level, phrase-level, and detailed multi-sentence localized image and video captioning.
\end{abstract}

\abscontent

\section{Introduction}
\label{sec:intro}
Image captioning has been a longstanding challenge in computer vision and natural language processing~\cite{chen2015microsoft}, as it involves understanding and describing visual content in natural language. 
While recent Vision-Language Models~(VLMs) have achieved impressive results in image-level captioning, generating detailed and accurate captions for specific regions within an image remains an open problem.
This challenge intensifies with videos, where models must additionally capture dynamic visual content, such as human actions, object motions, and human-object interactions. If resolved, it would open new doors for fine-grained grounded image/video understanding~\cite{zhong2022regionclip,liu2024grounding} and generation~\cite{li2021grounded,li2023gligen}.

Most existing VLMs (\eg, GPT-4o~\cite{gpt4o}) lack mechanisms for precise localization. %
Recent approaches that empower VLMs to take 2D localization cues such as bounding boxes~\citep{wang2023caption,wu2022grit,zhao2024controllable,huang2024segment} often yield brief phrases rather than detailed descriptions. While there are methods \cite{chen2023shikra,zhang2023gpt4roi,you2023ferret,yuan2024osprey,zhang2024omg,lin2024draw} that produce longer captions, they provide minimal detail or include unrelated content from other regions, as shown in~\cref{fig:comparison}. This raises the question: \textit{What makes detailed localized captioning (DLC) so challenging?}

\noindent{}We identify three key obstacles to DLC:
\begin{enumerate}
    \item \textbf{Loss of Region Details:} As shown in~\cref{fig:arch_comparison}, prior methods extract local features from global image representations, often leading to loss of fine-grained details—particularly for small objects in complex scenes. By the time the LLM processes the visual features, crucial details necessary for generating precise captions are already lost. Cropping the region of interest may enhance detail but risks losing essential contextual cues. %
    \item \textbf{Scarcity of High-Quality Datasets:} Datasets such as RefCOCOs~\cite{kazemzadeh2014referitgame,mao2016generation} and Visual Genome \cite{krishna2017visual} typically offer only short phrases that do not suffice for training models to generate rich, detailed captions. Recent synthetic data approaches \cite{you2023ferret,lin2024draw} are based on bounding boxes that could not precisely convey the exact region of interests, while methods that rely on global captions~\cite{guo2024regiongpt} may have difficulty capturing non-salient regions. %
    \item \textbf{Limitations in Benchmarks:} Prior localized captioning benchmarks compare generated captions against reference captions using language-based image captioning metrics~\cite{vedantam2015cider,banerjee2005meteor,papineni2002bleu,lin2004rouge,anderson2016spice} or LLM-based scoring~\cite{yuan2024osprey,yuan2024videorefer}. However, such techniques are not very applicable to DLC. Since the reference captions provided in the benchmarks often lack comprehensive details of the region, DLC models are often unfairly penalized for correct details not explicitly mentioned in the reference. %
\end{enumerate}

\noindent{}We propose the following solutions to these challenges:

To tackle the loss of details in regional features, we propose the \textbf{\underline{D}escribe \underline{A}nything \underline{M}odel} (\ModelName{}), which preserves both local detail and global context. \ModelName{} achieves this through two key innovations~(\cref{fig:architecture}): \textbf{1)} the \textit{focal prompt}, which encodes the region of interest with high token density, and \textbf{2)} the \textit{localized vision backbone}, which ensures precise localization while integrating global context. These components enable \ModelName{} to generate detailed and accurate captions, even for small objects in complex scenes.

To overcome the scarcity of high-quality DLC datasets, we introduce \textbf{\underline{S}emi-supervised learning (SSL)-based \underline{D}ata \underline{P}ipeline} (DLC-SDP) to generate high-quality localized captions in two stages. First, leveraging high-quality masks and keywords (\eg, class names, part names, or entities) from human-annotated segmentation datasets, we query a VLM to expand each keyword into a detailed caption given each mask-referred region. Second, inspired by self-training-based SSL in image classification~\citep{berthelot2019mixmatch,berthelot2019remixmatch,sohn2020fixmatch,xie2020unsupervised,lee2013pseudo}, DLC-SDP performs self-training with web images as an unlabeled dataset and segmentation datasets as labeled data. An LLM further summarizes descriptions into multiple granularities, yielding a diverse dataset with high-quality localized captions, enabling our model to outperform strong API-only baselines such as GPT-4o~\cite{gpt4o} and o1~\cite{o1}. %

To mitigate the limitations of current benchmarks, we introduce \textbf{DLC-Bench}, which evaluates detailed localized captions based on a set of predefined positive and negative attributes for each region, eliminating the reliance on comprehensive reference captions. This approach provides a more flexible and accurate evaluation, encouraging models to generate informative and precise descriptions.

\figArchComparison{t}
\tabAdvantages{t}

We summarize our contributions as follows:

\begin{enumerate}
    \item \textbf{Describe Anything Model (DAM):} A novel architecture with a focal prompt and a localized vision backbone for multi-granular regional image and video captioning.
    \item \textbf{SSL Data Pipeline (DLC-SDP):} A semi-supervised data pipeline that leverages high-quality segmentation annotations and unlabeled web images for scalable and diverse data curation.
    \item \textbf{DLC-Bench:} A benchmark designed to evaluate DLC without reference captions.
\end{enumerate}

Unlike generalist models, we focus on \textit{localized image and video captioning} across \textit{multiple granularities}, \textbf{achieving SOTA performance on 7 benchmarks} across keyword, phrase, and detailed multi-sentence captioning. Advantages of DAM, DLC-SDP, and DLC-Bench to prior practices are presented in \cref{tab:advantages}. We release our code, models, data, and benchmark at \href{https://describe-anything.github.io}{describe-anything.github.io}.

\section{Related Work}
\label{sec:related_work}

\noindent\textbf{Vision-Language Models (VLMs).} VLMs integrate visual and textual inputs for multimodal understanding and are broadly classified into BLIP-style~\citep{li2022blip,alayrac2022flamingo,awadalla2023openflamingo,li2023blip,dai2023instructblip,laurenccon2024obelics,dubey2024llama} and LLaVA-style~\citep{liu2024visual,liu2023improvedllava,liu2024llavanext,chen2024scaling,bavishi2023fuyu,zhu2023minigpt,bai2023qwen,wang2024qwen2,chen2023internvl,lin2024vila,beyer2024paligemma,wu2024vila}. However, these models lack precise localization capabilities, limiting their ability to generate regional descriptions.

\noindent\textbf{Localized Image Captioning.} While general VLMs generate image-level captions, localized captioning requires fine-grained regional descriptions. SoM~\citep{yang2023dawn,yang2023set} augments VLMs with visual markers, but these markers may blend with the background, as discussed in~\cref{sec:general_vlms}. Region-aware VLMs~\citep{chen2023shikra,yuan2024osprey,zhang2023gpt4roi,peng2023kosmos,wang2023all,rasheed2024glamm,sun2024alpha,guo2024regiongpt,lin2024draw,zhang2024omg,zhao2024controllable,wang2023caption,huang2024segment} introduce regional referring inputs. %
Recent efforts such as Merlin, Artemis, and VideoRefer~\citep{yu2024merlin,qiu2025artemis,yuan2024videorefer} extend region-based captioning to videos. However, these methods still struggle with capturing intricate details in the referring regions, as shown by the examples in~\cref{fig:comparison}. %
This is because prior models either extract localized features from global image embeddings or simply encode the referring condition as referring tokens, which leads to insufficient regional details for the LLM, especially for small objects. We address this via focal prompting and a localized vision backbone, balancing local detail with global context.

Another limitation is the scarcity of high-quality datasets. Datasets like RefCOCOs~\citep{kazemzadeh2014referitgame,mao2016generation} and VG~\citep{krishna2017visual} provide only short phrases. Recent approaches~\citep{you2023ferret,lin2024draw,guo2024regiongpt,rasheed2024glamm,yuan2024osprey,yuan2024videorefer} use bounding-box-based VLM queries, sometimes augmented with global captions, for synthetic generation, which leads to caveats discussed in~\cref{ssec:referring_boxes}. We propose an SSL data pipeline that uses human-annotated and unlabeled data for richer regional descriptions.

\noindent\textbf{Benchmarking Localized Captioning.} \cite{you2023ferret,zhang2024ferret,yuan2024osprey,guo2024regiongpt,rasheed2024glamm,zhang2024omg,huang2024segment,zhao2024controllable,qiu2025artemis} evaluate localized captioning by computing language-based image captioning metrics~\cite{vedantam2015cider,banerjee2005meteor,papineni2002bleu,lin2004rouge,anderson2016spice} between predicted captions and reference captions. However, these metrics focus on textual matching and may not correlate well with the factual correctness or quality for detailed descriptions. \cite{yuan2024osprey,yuan2024videorefer} use Sentence-BERT~\cite{reimers2019sentence} and text-only LLMs to score the predictions against reference captions. However, reference captions often lack comprehensive details about the region of interest, which penalizes models for correct details not explicitly mentioned in the reference by treating them as hallucinations, as discussed in~\cref{ssec:reference_captions}. Our DLC-Bench resolves this issue by eliminating the need for reference captions.

\noindent\textbf{Vision Models with Focus.} Prior works enhance attention to salient regions using focal self-attention~\citep{yang2021focal}, ToSA~\citep{singh2024tosa}, ToMe~\citep{bolya2022token}, DW-ViT~\citep{ren2022beyond}, Quadformer~\citep{ronen2023vision}, and V$*$~\citep{wu2023textit}. These methods allocate resources dynamically to salient regions defined by the model. In contrast, our focal prompt explicitly prioritizes user-specified regions, ensuring accurate and detailed captions even for non-salient objects.

\section{DAM: Describe Anything Model}
\label{sec:model}

\textbf{Describe Anything Model (\ModelName{})} generates detailed localized descriptions of user-specified regions within images and videos. \ModelName{} effectively balances local detail and contextual information through our proposed \textit{focal prompt} and \textit{localized vision backbone}.

\subsection{Task Formulation}

The task of detailed localized captioning involves generating comprehensive textual descriptions focused exclusively on specified regions within images or videos. Formally, given $N$ input frames $I^{(i)} \in \mathbb{R}^{H \times W \times 3}$ and corresponding binary masks $M^{(i)} \in {\{0,1\}}^{H \times W}$ indicating the region of interest in each frame, the objective is to produce a detailed description $T$ of the content within the region through a captioning model:
\begin{equation}
    T = \text{CaptioningModel}\left( \{ I^{(i)}, M^{(i)} \}_{i=1}^N \right)
\end{equation}

We focus on using binary masks $M^{(i)}$ as the localization input, since other forms of localization (\eg, points, scribbles, boxes, or masks on an image or a subset of frames in a video) can be transformed into masks via segmentation models such as SAM~\citep{kirillov2023segment} and SAM~2~\citep{ravi2024sam}. For simplicity, we first introduce our method for localized image captioning, omitting the frame index $i$, and later extend it to videos in~\cref{ssec:extension_to_videos}.

\subsection{Model Architecture}
\figArchitecture{t!}
As shown in \cref{fig:architecture}, \ModelName{} consists of two key components: the \textit{focal prompt} and the \textit{localized vision backbone}.

\subsubsection{Focal Prompt}

To provide a detailed representation of the region of interest within its context, we introduce the \textit{focal prompt}, which includes both the full image and a focal crop centered around the specified area, along with their corresponding masks.

We first extract the bounding box $B$ of the mask $M$ and expand it by a factor $\alpha$ in both the horizontal and vertical directions to include additional surrounding context:
\begin{equation}
    B' = \text{ExpandBox}(B, \alpha).
\end{equation}
For instance, setting $\alpha = 3$ results in a region that can be up to $9$ times as large as the original bounding box, subject to clipping at the image boundaries. If either the height or width of the expanded box is less than $48$ pixels, we enforce a minimum size of $48$ pixels in that dimension to ensure sufficient context for very small regions.

The focal crop of the image and mask are then:
\begin{equation}
    I' = I|{B'}, \quad M' = M|{B'},
\end{equation}
where $|{B'}$ denotes cropping to $B'$. The focal prompt thus consists of 1) the full image $I$ and its mask $M$ and 2) the focal crop $I'$ and its mask $M'$.
By including both the full image and the focal crop, along with their masks, the focal prompt contains both global context and a detailed view of the region of interest.

\subsubsection{Localized Vision Backbone}

Effectively processing all four components of the focal prompt with a VLM is non-trivial, as naively concatenating the full image and the focal crop leads to a loss in performance (\cref{tab:ablation_result}). 
We propose the \textit{localized vision backbone}, which 1) achieves localized understanding by encoding the masks in a spatially aligned manner and 2) integrates global context into the region of interest through gated cross-attention.

\noindent\textbf{Handling Localization Inputs.}
Similar to how an image is encoded by a linear patch embedding layer in vision transformers (ViTs)~\cite{dosovitskiy2020image}, we integrate the mask $M$ into its corresponding full image $I$ through another patch embedding layer that takes in 2D inputs with one channel. 

Specifically, the full image $I$ and its mask $M$ are processed through patch embedding layers, followed by the global vision encoder $f_{\text{G}}(\cdot)$ to obtain global visual features $\mathbf{z}$. The focal crop $I'$ and its mask $M'$ undergo a similar process with the regional vision encoder $f_{\text{R}}(\cdot)$, except that $f_{\text{R}}(\cdot)$ also takes $\mathbf{z}$ as a context to obtain the final fused visual features $\mathbf{z}'$. Specifically, we have:
\begin{equation}
    \mathbf{x} = E_{\text{I}} ( I ) + E_{\text{M}} ( M ) + P, \quad \mathbf{z} = f_{\text{G}} ( \mathbf{x} ),
\end{equation}
\begin{equation}
    \mathbf{x}' = E_{\text{I}} ( I' ) + E_{\text{M}} ( M' ) + P, \quad \mathbf{z}' = f_{\text{R}} ( \mathbf{x}', \mathbf{z} ),
\end{equation}
where $E_{\text{I}}(\cdot)$ and $E_{\text{M}}(\cdot)$ are the image and mask patch embedding layer, respectively, $\mathbf{x}$ and $\mathbf{x}'$ are global and focal embedded inputs with information for both the image and the mask, and $P$ denotes the positional encoding.

The newly added mask embedding layer $E_{\text{M}}$ is initialized to output zeros, ensuring that the VLM's initial behavior is unaffected prior to fine-tuning. %

\noindent\textbf{Regional Feature Encoding with Gated Cross-Attention Adapters.}
To integrate global context into the focal prompt, we insert gated cross-attention adapters~\citep{alayrac2022flamingo,li2022blip} into each transformer block of the regional vision encoder $f_{\text{R}}$. After the self-attention and feed-forward layers, we add a gated cross-attention mechanism that allows local features to attend to global features:
\begin{equation}
\mathbf{h}^{(l)'} = \mathbf{h}^{(l)} + \tanh \left( \gamma^{(l)} \right) \cdot \text{CrossAttn} \left( \mathbf{h}^{(l)}, \mathbf{z} \right),
\end{equation}
\begin{equation}
\mathbf{h}^{(l)}_\text{Adapter} = \mathbf{h}^{(l)'} + \tanh \left( \beta^{(l)} \right) \cdot \text{FFN} \left( \mathbf{h}^{(l)'} \right),
\end{equation}
where $\mathbf{h}^{(l)}$ is the output of the $l$-th self-attention block in $f_{\text{R}}$, $\gamma^{(l)}$ and $\beta^{(l)}$ are learnable scaling parameters initialized to zero, and $\text{CrossAttn}$ denotes cross-attention with queries from $\mathbf{h}^{(l)}$ and keys and values from the global features $\mathbf{z}$, similar to how cross-attention is employed in encoder-decoder Transformers~\cite{vaswani2017attention}. $\mathbf{h}^{(l)}_\text{Adapter}$ is used in place of $\mathbf{h}^{(l)}$ in the next Transformer block. To reduce the number of parameters, $f_{\text{R}}$ shares self-attention block weights with $f_{\text{G}}$.

By initializing $\gamma^{(l)}$ and $\beta^{(l)}$ to zero, we ensure that the initial behavior of the model remains identical to the original VLM prior to fine-tuning. During training, the model learns to leverage the global context to enhance local feature representations, facilitating detailed and contextually accurate descriptions.

\noindent\textbf{Generating Detailed Localized Descriptions.}
The visual features from both the global and regional vision encoders are combined and fed into the large language model to generate detailed, context-aware descriptions $T$:
\begin{equation}
T = \text{LLM} ( \mathbf{t}, \mathbf{z}' ),
\end{equation}
where $\mathbf{t}$ denotes textual prompt tokens.

Notably, the proposed components do \textit{not} increase the sequence length of the vision tokens, ensuring that \ModelName{} remains efficient. By initializing new modules (mask embedding $E_{\text{M}}$ and scaling parameters $\gamma^{(l)}$ and $\beta^{(l)}$) to zeros, we preserve the pre-trained capabilities of the VLM prior to fine-tuning, \textit{allowing for smooth adaptation of an off-the-shelf VLM without rerunning pre-training}. Thanks to this design, our model requires way less training data ($\sim\!1.5$M samples) than prior works that involve VLM pretraining. %

\subsection{Extension to Videos}
\label{ssec:extension_to_videos}

Since images can be considered as videos with a single frame, the model naturally extends to handling videos by processing sequences of frames and their corresponding masks. The visual features from all frames are concatenated in the sequence dimension and fed into the language model to generate detailed localized descriptions across the video frames, compatible with how VLMs are pretrained to handle videos. We leverage SAM 2~\citep{ravi2024sam} to turn sparse localizations into a mask for each frame.

\figBenchmark{t!}

\begin{table*}[t]
    \centering
    \begin{minipage}{0.32\textwidth}
        \centering
        \tabSemMetrics{t}
    \end{minipage}
    \hfill
    \begin{minipage}{0.30\textwidth}
        \centering
        \tabFlickr{t}
    \end{minipage}
    \hfill
    \begin{minipage}{0.36\textwidth}
        \centering
        \tabRef{t}
    \end{minipage}
\end{table*}

\section{DLC-SDP: SSL-based Data Pipeline}
\label{sec:pipeline}

The effectiveness of \ModelName{} depends critically on the availability of high-quality training data for detailed localized descriptions. %
To this end, we propose the \textbf{\underline{S}emi-supervised learning (SSL)-based \underline{D}ata \underline{P}ipeline} (DLC-SDP), a two-stage approach that enables us to build a large and diverse dataset with high-quality localized descriptions.

\subsection{Stage 1: Leveraging Existing Annotations}
The first stage of DLC-SDP reframes the data generation problem into a \textit{vision-grounded description expansion task}. We observed that although current VLMs often struggle to generate detailed localized descriptions when given a referring mask, they can effectively expand short localized descriptions into detailed ones. Inspired by this observation, we leverage high-quality human-annotated masks and keywords (object class names, part names, entities, etc.) from existing segmentation datasets, reframing the VLM query to expand each regional keyword into a detailed caption given the referring masks.

Importantly, our model is trained to predict these high-quality descriptions \textit{without} taking the initial keywords as inputs. Since there are no class labels provided at inference time--and existing VLMs perform poorly without them--our approach ensures superior data quality compared to direct VLM prompting for distillation.

\subsection{Stage 2: SSL with Unlabeled Data}
Since it is not scalable to rely on high-quality manual annotations, the second stage of DLC-SDP employs self-training-based semi-supervised learning techniques inspired by successful approaches in image classification~\citep{xie2020self,berthelot2019mixmatch,berthelot2019remixmatch,sohn2020fixmatch}. Our self-training approach involves four steps:

\begin{enumerate}
    \item \textbf{Mask Generation.} We use open-vocabulary segmentation models~\citep{lin2024vila,huang2024segment} to extract object masks from unlabaled web images.
    \item \textbf{Description Generation.} Our \ModelName{}, initially trained on the DLC dataset based on annotated segmentation datasets, generates detailed localized descriptions for these regions.
    \item \textbf{Confidence-based Filtering.} We apply CLIP-based confidence filtering to keep only high-quality samples, following SSL literature.
    \item \textbf{Data Expansion.} The newly generated (image, mask, description) triplets are added to our training dataset.
\end{enumerate}

This semi-supervised approach dramatically expands the range of object categories and increases data diversity beyond the initial supervised dataset. Furthermore, to support the capability of \ModelName{} for multi-granular captioning, we leverage an LLM to summarize the detailed descriptions into shorter forms (\eg, phrases or short sentences), enabling \ModelName{} to flexibly generate captions ranging from succinct phrases to multi-sentence narratives.

By systematically curating our training data through this two-stage SSL pipeline, \ModelName{} achieves significant performance improvements. Notably, our model trained on data obtained with DLC-SDP outperforms GPT-4o~\citep{gpt4o} and o1~\citep{o1}, two strong closed-sourced baselines, showing the effectiveness of DLC-SDP. We present implementation details of DLC-SDP in~\cref{sec:implementation_details}.

\section{DLC-Bench: Benchmark for DLC}
\label{sec:benchmark}
We introduce \textbf{DLC-Bench}, a benchmark designed for DLC to eliminate the need for comprehensive reference captions. The core intuition behind DLC-Bench is that an ideal description should be rich in relevant details while strictly avoiding factual errors or information for irrelevant regions. Therefore, we assess predictions based on a set of predefined positive and negative attributes for each region. 

As illustrated in \cref{fig:benchmark}, the evaluation process for a model like \ModelName{} has two steps:

\noindent{}\textbf{1.} The model is prompted to generate a detailed description for each masked region in the benchmark dataset. \\
\noindent{}\textbf{2.} An LLM serves as a judge, assessing the generated description by responding to a set of manually curated positive and negative questions about region details.

DLC-Bench employs two categories of questions for each annotated instance, with example in~\cref{fig:benchmark_example}:
\begin{itemize}
    \item \textbf{Positive questions} focus on specific attributes of object parts that \textit{should} be present in the description. The model earns a point if the description accurately includes the specified detail; omissions receive no points, while factual errors incur a penalty. %
    \item \textbf{Negative questions} focus on details that \textit{should not} be present--either attributes typical of similar objects but absent in the target instance, or descriptions irrelevant to the specified region. A point is awarded if the model correctly omits such details; conversely, including them results in a penalty. To avoid getting high scores for captions that are completely off, point could only be awarded if the caption has the correct recognition of the object.
\end{itemize}

This approach provides a more flexible and accurate evaluation, encouraging models to generate informative and precise descriptions without constraints from incomplete reference captions.

Our DLC-Bench comprises a total of 892 manually verified questions covering a wide range of attributes and potential cases for hallucinations. Details on the curation process and the scoring mechanism are provided in \cref{sec:dlc_bench_details}.

\section{Results}
\label{sec:experiments}

\ModelName{} excels at \textit{localized image and video captioning} across \textit{multiple granularities} including keyword, phrase, and detailed captions, \textbf{achieving SOTA on 7 in-domain and zero-shot benchmarks} (\cref{tab:keyword_lvis_paco,tab:flickr30k,tab:ref_l4,tab:main_result,tab:hc_stvg,tab:videorefer}). %
We explain the details for each benchmark in~\cref{sec:eval_setup}.

\subsection{Quantitative Results}
\label{sec:quantitative_results}

\noindent\textbf{Open-class keyword-level localized captioning} task requires the model to output keywords containing the object and part entities to describe the region. Tested on object-level LVIS~\cite{gupta2019lvis} and part-level PACO~\cite{ramanathan2023paco} datasets in \cref{tab:keyword_lvis_paco}, our method achieves state-of-the-art performance. In the PACO benchmark, a challenging benchmark that includes both full objects and parts in complex scenes and requires the model to decide whether the region is an object or a part, our method achieves 73.2\% semantic IoU and 84.2\% semantic similarity, outperforming the previous best by 23.2\% and 8.5\% respectively.

\noindent\textbf{Phrase-level localized captioning} task requires the model to output a phrase containing a brief description for each region that includes object identification and attributes typically within a few words. Tested zero-shot on Flickr30k Entities~\cite{plummer2015flickr30k}, our model achieves strong performance, outperforming the previous best by 12.3\% relative improvement on an average of 5 metrics.

\noindent\textbf{Detailed localized captioning} task requires the model to output a detailed description for each the region with the length spanning from a long sentence to multiple sentences. An ideal description includes the description of the object in the region, its parts, as well as their attributes and relationships. We benchmark this capability on the challenging Ref-L4~\cite{chen2024revisiting} benchmark and our proposed DLC-Bench. On the Ref-L4 benchmark in \cref{tab:ref_l4}, our method achieves 33.4\% relative improvement on average over the previous best on short language-based captioning metrics~\cite{vedantam2015cider,banerjee2005meteor,papineni2002bleu,lin2004rouge,anderson2016spice}, and 13.1\% relative improvement on the long language-based captioning metrics~\cite{chan2023clair}. %

We also benchmark various regional captioning models on \textbf{our proposed DLC-Bench}, which does not suffer from the limitations of requiring reference captions in previous benchmarks. As shown in \cref{tab:main_result}, our Describe Anything Model (\ModelName{}) significantly outperforms existing general and region-specific VLMs, achieving state-of-the-art positive and negative accuracy and demonstrating its ability to produce detailed and accurate descriptions. Remarkably, DAM surpasses GPT-4o~\citep{gpt4o} and o1~\citep{o1}, two strong API-only baselines. DAM also surpasses models with thinking mode enabled~\cite{o1,team2025claude,team2025gemini}. %

\noindent\textbf{Detailed localized video captioning} requires the model to output a detailed description for each region in a video. We benchmark this capability on the challenging HC-STVG~\cite{tang2021human} benchmark and the detailed captioning benchmark proposed by VideoRefer~\cite{yuan2024videorefer}. In \cref{tab:hc_stvg}, our proposed \ModelName{} achieves 19.8\% relative improvement over the previous best on HC-STVG, including concurrent work VideoRefer~\cite{yuan2024videorefer}. In \cref{tab:videorefer}, the benchmark proposed by concurrent work VideoRefer~\cite{yuan2024videorefer}, our proposed \ModelName{} surpasses the previous best in \textit{both zero-shot and in-domain settings}, where zero-shot indicates not being trained on in-domain datasets derived from Panda-70M~\cite{chen2024panda}, which the benchmark also sources videos from.

Finally, we analyzed the performance of \ModelName{} in HD (hallucination detection) sub-task and found that \ModelName{} often predicts correct details not present in the reference caption. This indicates that the lower zero-shot performance on this sub-task is \textit{not necessarily due to the hallucination of our model} but rather due to the missing details in the reference caption. %
We illustrate this further in~\cref{ssec:reference_captions}. %

\tabMainResult{t!}
\tabHCSTVG{t!}

\subsection{Qualitative Results}

Qualitative comparisons in~\cref{fig:comparison} show that \ModelName{} excels in both accuracy and the level of details. %

\noindent\textbf{Detailed Localized \textit{Video} Captioning.} DAM describes user-specified objects in videos with localization from any frame. As shown in \cref{fig:video_examples} \textbf{(a)}, DAM effectively describes objects under challenging conditions, such as motion and occlusion. \textit{We offer more video examples in~\cref{ssec:additional_video_examples}.}

\noindent\textbf{Controlling Description Granularity.} As shown in \cref{fig:control}, \ModelName{} allows control over the amount of details and length of descriptions with different prompts.

\noindent\textbf{Zero-shot 3D Object Captioning.} Our model can also describe objects in multi-view datasets such as Co3Dv2~\cite{reizenstein21co3d}, integrating information from multiple frames to provide coherent descriptions of 3D objects (\cref{fig:video_examples}(b)). %

\subsection{Ablations}
\noindent\textbf{Visual Prompting.} We analyze different prompting strategies and find that both localized inputs and contextual information are crucial for accurate descriptions. Using only the full image limits focus on specific regions (48.7\%), while local crops improve detail but lose context (60.1\%). Simply concatenating both performs poorly (42.4\%). Adding cross-attention significantly improves performance (63.2\%), and using focal crops further enhances results (65.4\%). Our best approach, the \textbf{focal prompt}, integrates focal crops with cross-attention, achieving \textbf{67.3\% accuracy} without increasing sequence length for the LLM. 

\tabVideoRefer{t!}
\figComparison{t!}
\figVideoExamples{t!}
\figControl{t!}

\tabAblationResult{t!}
\tabAblationData{t!}

\noindent\textbf{Data Scaling.}
Expanding supervised datasets boosts performance, demonstrating the value of diverse regional data. Incorporating \textbf{semi-supervised learning (SSL)} with 10\% of unannotated SA-1B images further improves accuracy to \textbf{67.3\%}, showcasing our data pipeline’s scalability.

Additional ablations, including training a model with the architecture from prior work on our data, comparing image-only vs image-video joint training, and ablating prompt augmentations, are in \cref{sec:additional_ablation_studies}.

\section{Discussions and Conclusion}
\label{sec:conclusion}

We introduced \ModelName{}, a model for detailed localized captioning in images and videos, balancing local detail and global context through a focal prompt and localized vision backbone. We developed DLC-SDP, an SSL data pipeline leveraging segmentation datasets and unlabeled web images for high-quality captions. We also proposed DLC-Bench, a benchmark using attribute-based evaluation to overcome limitations of reference-based scoring. \ModelName{} achieves SOTA in 7 benchmarks in multi-granular regional captioning. We present additional discussions in~\cref{sec:discussions}.

\clearpage
\setcounter{page}{1}
\renewcommand{\thefigure}{A.\arabic{figure}}
\setcounter{figure}{0}
\renewcommand{\thetable}{A.\arabic{table}}
\setcounter{table}{0}
\appendix

\figLocalizedInputs{h!}

\section{Challenges in Generating Detailed Localized Descriptions with Off-the-Shelf VLMs}
\label{sec:general_vlms}
Although cutting-edge Vision-Language Models (VLMs), such as GPT-4o~\cite{gpt4o} and LLaVA~\cite{liu2024visual,liu2024llavanext,liu2023improvedllava}, excel at generating global-level image descriptions, producing detailed \textit{localized} image captions remains an open problem. Specifically, these VLMs only take in RGB images along with text prompts and do not allow users to accurately specify regions of interest.

While users could employ text to localize the object to be described, this approach is often cumbersome and inefficient, requiring precise referring phrases that may still be difficult for the VLM to interpret. This can lead to mislocalization of the intended object, as illustrated in \cref{fig:localized_inputs}(a).

The required effort for both the user and the model can be significantly reduced if the user is allowed to specify the region directly using a representation in 2D coordinates that the model can understand. With this idea in mind, we focus on \textit{generating detailed localized descriptions} by enabling users to specify a region in an image for the model to describe in detail. Since spatial representations such as points and boxes can be converted into masks using SAM~\cite{kirillov2023segment} and SAM 2~\cite{ravi2024sam}, we concentrate on regions specified by mask inputs.

A first attempt to address this problem with existing VLMs is to reduce the task to global image captioning by presenting only the region to the VLM through masking or cropping, as shown in \cref{fig:localized_inputs}(b). While this forces the VLM to focus solely on the specified region, freeing users from the burden of expressing localizations as phrases, the lack of contextual information makes the task much more challenging and often confuses the VLM. This confusion can prevent the model from correctly identifying the object, let alone providing detailed descriptions of its parts. In more extreme cases, the model may even refuse to caption the region due to insufficient information in the cropped or masked image. Therefore, generating detailed localized captions requires more than just the local region.

An alternative approach to prompt existing off-the-shelf VLMs for localized descriptions is to overlay markings such as points, scribbles, contours, and alpha masks on the image~\citep{yang2023dawn, yang2023set}, as shown in \cref{fig:localized_inputs}(c). However, these markings may blend into the object or the background in highly complex scenes, making them unrecognizable to the VLMs. This issue is especially common for small objects that are not the main focus of the scene. Furthermore, the markings may render the image out-of-distribution, confusing the VLMs and disrupting the quality of output that they were originally capable of generating.

The exploration above highlights a conflict between the precision of localization and the availability of context. On one hand, we want the model to accurately focus on a specific region without mentioning other regions, such as other objects or the background. On the other hand, the model needs to leverage contextual information to correctly identify the object in the region of interest. This conflict makes it very difficult for current VLMs to produce high-quality localized descriptions.

Our proposed model overcomes this challenge by taking the localization as a \textit{separate} input in 2D space. This approach has the advantage of making the localization more explicit for the VLMs to parse while keeping the image within its original distribution, thus preventing the model from being distracted by the markings. This technique leads to accurate localization even in complex scenes, as illustrated in \cref{fig:localized_inputs}(d). Note that since \cref{fig:localized_inputs}(d) mainly focuses on explaining the design choices of inputting mask inputs to the model, focal prompting is included as a part of the model and is omitted in this figure for simplicity. We refer readers to \cref{fig:architecture} for illustrations on focal prompting.

\section{Evaluation Benchmarks}
\label{sec:eval_setup}

Our \ModelName{} is designed to perform well at \textit{localized image and video captioning} across \textit{multiple granularities}, including keyword, phrase, and detailed captions. Therefore, we evaluate and achieve SOTA in 7 in-domain and zero-shot benchmarks:

\begin{enumerate}
    \item The LVIS open-class keyword-level benchmark in ~\cref{tab:keyword_lvis_paco}.
    \item PACO open-class keyword-level benchmark (including object and parts as regions) in~\cref{tab:keyword_lvis_paco}.
    \item Flickr30k Entities phrase-level benchmark in~\cref{tab:flickr30k}.
    \item Ref-L4 detailed captioning benchmark in~\cref{tab:ref_l4}.
    \item Our proposed DLC-Bench detailed localized captioning benchmark in~\cref{tab:main_result}.
    \item HC-STVG detailed video captioning benchmark in~\cref{tab:hc_stvg}.
    \item VideoRefer detailed video captioning benchmark in~\cref{tab:videorefer}.
\end{enumerate}

We offer an explanation for each setup.

\subsection{Keyword-level Localized Captioning Benchmarks}
Open-class keyword-level localized captioning benchmarks, proposed in~\cite{yuan2024osprey}, require the model to output keywords containing the object and part entities to describe the region. In contrast to closed-class keyword-level localized captioning, which constraints the model output to several choices provided, open-class keyword-level localized captioning takes free-form text outputs from the model. The evaluation results are in~\cref{tab:keyword_lvis_paco}.
\begin{enumerate}
    \item For LVIS~\cite{gupta2019lvis}, this involves predicting the class name as a keyword, given the segmentation mask of an object. A typical class name ranges from one word to four words.
    \item For PACO~\cite{ramanathan2023paco}, this involves predicting the class name of an object in the mask if the mask contains a full object, or the object name and the part name if the mask contains an object part. This is especially challenging because it would require the model to understand nuances between full objects and object parts.
\end{enumerate}

\subsection{Phrase-level Localized Captioning \\Benchmarks}
Phrase-level localized captioning task requires the model to output a phrase containing a brief description for each the region that includes object identification and attributes typically within a few words. The metrics typically used in phrase-level benchmarks are CIDER, METEOR, BLEU, ROUGE\_L, and SPICE \cite{vedantam2015cider,banerjee2005meteor,papineni2002bleu,lin2004rouge,anderson2016spice}. We refer these metrics as short captioning metrics, as opposed to metrics from LLM-based evaluations that support evaluating detailed captions.

    We perform zero-shot evaluation on the grounded phrases in Flickr30k Entities~\cite{plummer2015flickr30k}, where our model is not trained on the entities annotated in the training split of Flickr30k Entities. Results are in~\cref{tab:flickr30k}.

\subsection{Detailed Localized Captioning Benchmarks}  
Detailed localized captioning task requires the model to output a detailed description for each the region with the length spanning from a long sentence to multiple sentences.

\begin{enumerate}
    \item We perform zero-shot evaluation on detailed captions in the Objects365~\cite{shao2019objects365} split of Ref-L4~\cite{chen2024revisiting} since we do not train on Objects365 dataset. We evaluate the prediction quality by computing short captioning metrics and CLAIR~\cite{chan2023clair} score against the reference captions in the dataset. We use CLAIR to evaluate raw detailed outputs, while we summarize both the prediction and ground truth with GPT-4o-mini~\cite{gpt4o} before evaluation with short captioning metrics. No ground truth or reference captions are provided to GPT-4o-mini, with the LLM setting exactly the same for all models for fairness. Results are in~\cref{tab:ref_l4}.
    \item We evaluate our model with DLC-Bench, our proposed benchmark for fine-grained region-based captioning. This evaluation is also zero-shot. We present details about our benchmark in~\cref{sec:dlc_bench_details}. Results are in~\cref{tab:main_result}.
\end{enumerate}

\subsection{Detailed Localized Video Captioning Benchmarks}
\begin{enumerate}
    \item We conduct evaluation on HC-STVG~\cite{tang2021human}, a spatial-temporal video grounding dataset with detailed captions used in prior and concurrent work~\cite{qiu2025artemis, yuan2024videorefer}. Following prior work~\cite{qiu2025artemis}, we evaluate the quality of localized captions with CIDER, METEOR, BLEU, ROUGE\_L, and SPICE \cite{vedantam2015cider,banerjee2005meteor,papineni2002bleu,lin2004rouge,anderson2016spice}. Results are in~\cref{tab:hc_stvg}.
    \item We also perform evaluation on the detailed localized video description benchmark in VideoRefer-Bench proposed by concurrent work~\cite{yuan2024videorefer}. GPT-4o is used to provide four dimensions of scores on a scale of 1 to 5. The four dimensions are Subject Correspondence (SC), Appearance Description (AD), Temporal Description (TD), and Hallucination Detection (HD). Zero-shot setting indicates that our model is not trained on Panda-70M~\cite{chen2024panda}, the dataset that VideoRefer-Bench sources the videos from. In-domain setting indicates mixing the detailed caption subset of VideoRefer-700k, which is also curated from Panda-70M~\cite{chen2024panda}, into our training data. Results are in~\cref{tab:videorefer}.
\end{enumerate}

\section{Discussions}
\label{sec:discussions}

\figMaskRoI{t!}
\figHallucinationExamples{t!}
\subsection{The Caveats of Using Referring Boxes in Data Pipeline}
\label{ssec:referring_boxes}
Caveats exist when boxes are used to refer to regions in the data pipeline. As shown in~\cref{fig:mask_roi}, boxes can be ambiguous in terms of what they are referring to, causing uncertainty for the VLM that we use in our data pipeline. In contrast, masks are much more specific in terms of the region that it is referring to. This motivates us to use manually annotated masks in existing segmentation datasets rather than bounding boxes in order to curate high-quality data for DLC with little referring ambiguity. We additionally take in manually annotated keywords (\eg, class names, part names, entities) in the datasets for regions we are annotating in our data pipeline to further reduce the ambiguity and potential confusion for our VLM in the data pipeline.

\subsection{The Pitfall of Using Reference Captions in Benchmarks}
\label{ssec:reference_captions}
As discussed in~\cref{sec:benchmark,sec:quantitative_results}, caveats exist for using a ``ground truth'' reference caption for benchmarking localized descriptions. Specifically, since such a reference caption is hardly comprehensive and may not contain all the details about the region of interest, the metrics from the benchmark will treat the correct details in the caption prediction about the region of interest that are not mentioned in the ground truth reference caption as hallucinations. This discourages the model from generating detailed captions.

We analyzed the performance of our method in HD (hallucination detection) sub-task in VideoRefer-Bench~\cite{yuan2024videorefer} and found that our model often predicts correct details that are not present in the reference caption. Specifically, the example in~\cref{fig:hallucination_examples} shows this phenomenon. While our model's prediction includes appearance and motion details about the change of the person's gesture and expression, such details are not mentioned in the reference caption in the dataset. Since the GPT evaluator does not see the video and uses the ground truth caption as the only source of information, it incorrectly believes that the gestures and expressions are hallucinations and gives our caption a very low score for the hallucination detection dimension. However, the evaluation is not valid, as our model is correct in the descriptions about the gestures and expressions. 

This indicates that the lower score on this sub-task is \textit{not due to the hallucination of our model}, but rather due to the missing details in the reference caption and the fact that our model, evaluated in a zero-shot setting, does not have awareness for what types of details are preferred by or included in the reference caption.

\figFailureCase{t!}
\subsection{Failure Cases}
We show two failure cases of \ModelName{} in \cref{fig:failure_case}. In \cref{fig:failure_case}(a), \ModelName{} misrecognizes the frog-shaped slipper to be a frog. In \cref{fig:failure_case}(b), \ModelName{} describes the person as pulling the body upward. We expect these errors to be mitigated by broader data coverage.

\subsection{Potential Limitations}
\label{sec:limitations}
\ModelName{} is only trained for multi-granular localized captioning, especially for detailed localized captioning (DLC) and is not specifically optimized for other general vision-language tasks. However, \ModelName{} is designed for in-depth analysis for the task of multi-granular image and video localized descriptions rather than for breadth for general vision-language understanding, which justifies the design choice.

\subsection{Computational Efficiency}
\label{ssec:efficiency}
\ModelName{} incorporates our proposed localized vision encoder, which differs from the SigLIP~\cite{zhai2023sigmoid} vision encoder used in \cite{lin2024vila} by adding two key components: \textit{patch embedding layers} for encoding the mask and \textit{cross-attention blocks}. Importantly, these components do not alter the dimensions or sequence length of the vision features passed to the large language model, ensuring that the parameter count and computational efficiency of the large language model are unaffected. Since the vision encoder represents only a small fraction of the total parameters and computational operations, the overall increase in FLOPs and parameter count remains marginal, maintaining the model's efficiency.

To be more specific, unlike prior works that derive regional features from \textit{image features} for each region, the regional feature used in our approach comes directly from \textit{a global and a focal view of the input image}, with cross-attention enhancing the focal representation. This design is justified as the vision encoder is much smaller than the LLM (400M vs. 3B/8B parameters), with minimal latency impact (\textit{0.06s compared to 1.49s for 3B LLM} as measured in our pipeline). This overhead is outweighed by the benefits of preserving fine details that global image features miss as indicated in \cref{tab:ablation_result}), especially for small regions. Finally, \ModelName{} 3B outperforms much larger models in challenging (\cref{tab:main_result}), showing our efficiency.

\subsection{Training Data}
\label{ssec:training_data}
In addition to the details in data annotation presented in~\cref{ssec:data_annotation}, we discuss the training data of our work in this section and present a comparison with recent works. Compared with recent work Ferret \cite{you2023ferret} which used 1.1M \textit{unreleased} samples and RegionGPT \cite{guo2024regiongpt} which used 1.5M \textit{unreleased} samples, we train our model on a comparable amount of data (1.5M samples). However, we obtain much better performance~(\cref{tab:main_result}), which shows the effectiveness of DAM.

\subsection{Performances of Baseline Models on DLC-Bench}

Interestingly, region-specific VLMs often perform on par or worse than generic VLMs. This is likely because many are trained on datasets with short regional captions, leading them to produce brief, phrase-level descriptions. Even when prompted for longer descriptions~\cite{guo2024regiongpt,zhang2024omg}, these models tend to include irrelevant details about the background, speculations, or hallucinations, due to insufficient regional information. Providing crops instead of full images leads to mixed results for different region-specific VLMs since these models are not designed to describe regions in crops.

\figBenchmarkExample{t!}

\section{Details for DLC-Bench}
\label{sec:dlc_bench_details}
\noindent\textbf{Image and Instance Selection.} We leveraged a subset of the Objects365 v2~\cite{shao2019objects365} validation set, which was manually annotated with segmentation masks in~\cite{gu2024dataseg}, for image and instance selection. We collected a set of 892 challenging questions from this subset, each containing one object of interest. Each question is manually inspected, and questions with ambiguous or unclear answers are filtered out. To maintain the integrity of our benchmark, we conducted de-duplication to ensure that no images used in the benchmark were present in our training dataset for detailed localized captioning.

\noindent\textbf{Positive Question Generation.} For each masked region, we prompted an off-the-shelf Visual Language Model (VLM) to generate a list of parts. Subsequently, for the whole object and each part, we asked the VLM to generate a list of properties covering aspects such as color, shape, texture, materials, and size. Each property is stored in the form \texttt{([object name], [part name], [property name], [property value])}. For example, if the masked region is a corgi, the VLM could describe the brown fur of the corgi as \texttt{(corgi, fur, color, brown)}.

We used this list of properties as a starting point for manual curation. We then manually added significant properties that the VLM missed, revised inaccurate properties, and removed hallucinated or ambiguous properties from the VLM outputs. Finally, we turned these properties into questions that test whether a description accurately covers the property.

\noindent\textbf{Negative Question Generation.} We targeted mislocalization and hallucination, which are two types of negatives (\ie, cases in which a property or an object should not be included in the descriptions). Specifically, for mislocalization errors, we prompted the VLMs to generate a list of objects in the image that are not in the masked region. We also prompted the VLMs to generate a list of parts that are commonly associated with the object type of the masked region but are not present or visible in the masked object in the image (\eg, the head of a corgi if it is occluded and thus not included in the masked region).

To avoid biasing towards one specific off-the-shelf VLM, we leveraged multiple VLMs for different instances to generate initial positives and negatives. Specifically, we annotated 34 regions using GPT-4o~\cite{gpt4o}, 35 using Gemini 1.5 Pro~\cite{team2023gemini,team2024gemini}, and 31 using Anthropic Claude 3.5 Sonnet~\cite{team2024claude} for the initial property generation. We used the same image prompting method for all VLMs as we did when prompting the VLMs in the first stage of our data pipeline.

Note that the choices for each question are mutually exclusive, which ensures one option is always valid and leaves no room for two options to be true at the same time.

\noindent\textbf{Scoring Mechanism.} Our evaluation methodology involves scoring the models based on their ability to include correct details and exclude incorrect or irrelevant information.

To evaluate a model like DAM for its ability to output detailed localized captions, we first prompt the model to generate descriptions for each of the masked instances. Then, instead of directly asking our model to provide answers to these questions, we prompt a text-only LLM, Llama 3.1 8B~\cite{dubey2024llama}, to serve as a judge to rate the localized descriptions according to the positive and negative questions.

For each model-generated description, we apply the following scoring rules:

\begin{itemize}
\item \textbf{Positive Scoring:} For each positive question, if the description correctly includes the specified detail, the model receives a point. To prevent models from artificially inflating their scores by generating excessively long descriptions and guessing details, we penalize incorrect details and discourage models from including uncertain or erroneous content. If the detail is mentioned but incorrectly~(\eg, wrong color), a penalty of one point is applied. No point is awarded if the description does not mention the detail. Partial points (0.5 points) are awarded for answers that are partially correct but insufficiently detailed. Note that the model gets positive points only when the object recognition is correct, as the correctness of the details depends on the correctness of the overall region recognition. We present a positive example in~\cref{fig:benchmark_example}(c).
\item \textbf{Negative Scoring:} For each negative question, if the description appropriately excludes the incorrect or irrelevant detail, the model gets a point. If the description includes the detail, indicating mislocalization or hallucination, a penalty is applied. The model gets zero or negative points when the object recognition is incorrect, since otherwise a caption that is random and completely off could get high scores on the negative questions. We present a negative example in~\cref{fig:benchmark_example}(d).
\end{itemize}

The positive (negative) score for a model is the sum of points for positive (negative) questions, normalized by the maximum possible score to yield a percentage for comparison. We also average the positive and negative scores to obtain an overall score, which represents the model's overall capability in detailed localized captioning.

We present an example from DLC-Bench in \cref{fig:benchmark_example}. The example region in \cref{fig:benchmark_example}(a) features a stove with coil burners. An example description of the region is presented in \cref{fig:benchmark_example}(b). For the example positive question in \cref{fig:benchmark_example}(c), the LLM judge selects option C, as the caption correctly mentions that the control panel is at the back, allowing the model to get a point for this positive question. For the negative question in \cref{fig:benchmark_example}(d), the LLM judge selects option B, as the caption correctly indicates that it is not an induction cooktop, allowing the model to get a point for this negative question.

\noindent\textbf{Evaluation Setting.} For our models, we follow our inference setting described in~\cref{sec:implementation_details}.

\tabAblationModel{t!}
\tabAblationPromptAug{t!}
\tabAblationJointTraining{t!}

\section{Additional Ablation Studies}
\label{sec:additional_ablation_studies}

\noindent\textbf{Model Architecture with the Same Training Data.}
A model's performance is largely due to two factors: model architecture design and training data. Since both factors differ for different models, it is hard to compare the effectiveness of different model architectures head-to-head.

To this end, we compare our model architecture against VP-SPHINX~\cite{lin2024draw}, the strongest prior baseline in most benchmarks that we tested on. By continuously training a VP-SPHINX model~\cite{lin2024draw} on our data after pre-training on the originally proposed datasets. This is a fair comparison since our method is also fine-tuned from a pretrained VLM, VILA-1.5~\cite{lin2024vila}, with two stages of training prior to training on our region-specific dataset.

As shown in~\cref{tab:ablation_model}, our model architecture achieves much better performance on detailed localized captioning benchmark DLC-Bench with trained on the same dataset from our proposed data pipeline. This justifies that our proposed focal prompt and localized visual backbone are able to provide more detailed features compared to just the global image features extracted by a vision encoder on the full image with a regional referring feature, as employed in~\cite{lin2024draw}.

\noindent\textbf{Prompt Augmentation.}
We compared variants of our model with and without prompt augmentation. As shown in~\cref{tab:ablation_prompt_aug}, incorporating prompt augmentation slightly degrades our model's performance on the positive questions in our benchmark. We hypothesize that despite introducing variations in the prompts and enhancing the model's instruction-following capabilities, prompt augmentation creates a mismatch between the prompts used during training and those used during evaluation (as we always use the same prompt for evaluation, which is detailed in~\cref{ssec:inference_setting}). Since the prompt used during evaluation might not be the same as the prompt used in training, the model may also occasionally reference other tasks from our mixing dataset ShareGPT-4V for the length of outputs. This may cause the model to produce outputs that are not as detailed as when it is trained exclusively with the original prompt. Importantly, the model's performance on the negative questions remains unchanged, indicating that prompt augmentation does not lead to hallucinations or mislocalization.

Despite the slight degradation of the performance in the benchmark (0.6\% in the overall accuracy), we observed that prompt augmentation improves instruction-following capabilities when prompts include additional instructions, particularly those specifying requirements on the length of the outputs. Therefore, we default to using the model without prompt augmentation in our benchmark evaluations, including ablations. In contrast, we employ the model with prompt augmentation in the qualitative evaluations.

\noindent\textbf{Image-only Training vs Image+Video Joint Training.}
We also compared our image-only \ModelName{} with \ModelName{} with image + video joint training in~\cref{tab:ablation_joint_training}. We show that our model with image-video joint training slightly outperforms our model with image-only training on detailed localized image captioning. Note that for this ablation study, we keep the model size the same and use the 3B model for both image-only training and image-video joint training. We use image-only training as the default option for results in our benchmark for simplicity.

\tabSOMResult{t!}
\figAdditionalQA{t!}

\figAdditionalImageExamples{t!}
\section{Additional Quantitative Results}
\label{sec:additional_quantitative_results}
\noindent\textbf{Set-of-Marks Prompting.}
We present a comparison with baseline VLMs that use Set-of-Marks (SoM) prompting~\cite{yang2023set} in \cref{tab:som_result}. SoM leads to degraded results compared to the prompt engineering method used in stage one of our data annotation pipeline. This is mostly because the marks proposed by SoM blend in with the object or the background in complex scenes. They might also mask out some part of the object, which interferes with the model's understanding capabilities. Therefore, for fair comparisons, we use the same prompt engineering method as we use in stage one of our data annotation pipeline in our main result in \cref{tab:main_result}. Importantly, region-specific VLMs, including \ModelName{}, have predefined ways of encoding regional inputs, making SoM inapplicable to these models.

\section{Additional Qualitative Results}
\label{sec:additional_qualitative_results}

\subsection{Detailed Localized Image Captioning}
In~\cref{fig:additional_image_examples}, we present additional examples from LVIS~\cite{gupta2019lvis} to show our model's strong performance on detailed localized image captioning.

Our model demonstrates robust localization and region understanding capabilities. In the first example, it accurately describes the sofa cushion without mentioning the dog that is outside the masked region. In the second example, it correctly identifies the roller blind, which would be challenging to recognize based solely on a local crop without context. In the third example, the model provides a detailed description of the giraffe without referencing the birds perched on it, as they fall outside the masked region. These examples highlight our model's precise localization abilities and its effectiveness in perceiving regional details with contextual understanding.

\subsection{Zero-shot QA Capabilities}
Although not trained on any regional QA datasets, \ModelName{} surprisingly exhibits emerging zero-shot capabilities on regional QA. 

In~\cref{fig:qa2}, we show examples of our model performing zero-shot QA. \ModelName{} is able to identify properties of objects in the masked regions. For example, it is able to identify the color of the clothing, the material of the stick, and the textural pattern of the fish in the first three examples. \ModelName{} is also capable of performing object recognition for a region in the image, identifying the strawberry in the last image.

\figAdditionalVideoExamples{t!}
\figAdditionalVideoExamplesPartTwo{t!}

\subsection{Detailed Localized Video Captioning}
\label{ssec:additional_video_examples}
We present more examples for detailed localized video captioning in~\cref{fig:additional_video_examples} and \cref{fig:additional_video_examples_part2}. Our model can describe objects in videos with large object motion and camera motion. \ModelName{} can also identify stationary objects by indicating that they are stationary in the description.

\figComparisonWithGPTImage{t!}
\figComparisonWithGPTVideo{p!}
\subsection{Qualitative Comparisons with Strong Baselines}
\noindent\textbf{Detailed Localized Image Captioning.} We also present qualitative comparisons with GPT-4o~\cite{gpt4o} and our strongest open-weight baseline VP-SPHINX~\cite{lin2024draw} in detailed localized image captioning in \cref{fig:comparison_with_gpt4o_image}.

In both examples, GPT-4o could not correctly recognize the objects in the masked regions, providing only vague descriptions. VP-SPHINX, while better than GPT-4o, still struggles with accurate object recognition and detailed descriptions. In the left image, VP-SPHINX incorrectly describes a group of seals when the masked region contains only one seal. In the right image, VP-SPHINX identifies the towel but provides minimal detail, missing key attributes like its color and texture.

Our model outputs detailed and high-quality descriptions of the seal and the towel. This improvement stems from our model's design which enables the fusion of object-specific information with broader contextual understanding.

\noindent\textbf{Detailed Localized Video Captioning.} We present comparisons with three strong video understanding models, GPT-4o~\cite{gpt4o}, Qwen2.5-VL~\cite{Qwen2.5-VL}, and recent work VideoRefer-~\cite{yuan2024videorefer}, in detailed localized video captioning in \cref{fig:comparison_with_gpt4o_video}. In the top example, we observed that GPT-4o struggles to interpret the cow's movements. Similarly, Qwen2.5-VL-7B incorrectly perceives the cow as stationary. VideoRefer-7B provides minimal motion and appearance details. In contrast, our 8B model accurately identifies the motion of the cow, providing more detailed information about it.

In the bottom example, GPT-4o misidentifies the object, mistakenly assuming the animal is transforming into a wolf or a pig. Meanwhile, Qwen2.5-VL-7B believes only the sheep's head is moving. VideoRefer-7B recognizes that the sheep is moving but provides little detail about the appearance of the sheep. In contrast, our model correctly identifies the animal in the masked region as a sheep throughout the video and accurately recognizes its full movement, providing details about its motion and appearance.

\tabDatasets{t!}
\tabVideoDatasets{t!}

\section{Implementation Details}
\label{sec:implementation_details}
\subsection{Data Annotation Pipeline}
\label{ssec:data_annotation}
\noindent\textbf{Stage 1.} We annotate four existing instance and semantic segmentation datasets for detailed localized descriptions. We use off-the-shelf VLMs for region annotations, with 603k
regions across 202k images with detailed localized descriptions in total in stage 1. For the model variant used in PACO~\cite{ramanathan2023paco} open-class dataset evaluation, we additionally merged in 81k annotated instances from PACO~\cite{ramanathan2023paco} to improve its part description capabilities, leading to 684k annotated regions, as detailed in~\cref{tab:datasets}. To prompt a VLM to output detailed localized descriptions, we input a cropped image and a masked image. While the cropped image allows coarse localization and provides high token density per pixel for clear descriptions, the masked image helps localize the object of interest when there are multiple instances with the same category. The category name is also provided in the text prompt, relieving the model from having to identify the object without the context from the image. We present the prompt for data annotation in~\cref{tab:prompt_for_data_annotation}.

\noindent\textbf{Stage 2.} We annotate 10\% of SA-1B through self-labeling, resulting in 774k annotations across 593k images, as detailed in~\cref{tab:datasets}. Due to filtering, the final number of instances and images is lower than the original 10\% subset of SA-1B. We do not use the masks provided with SA-1B, as they contain a large number of masks for parts. Instead, we employ the open-vocabulary detector OWL-ViT v2~\cite{minderer2022simple,minderer2024scaling} to detect objects in the images, and then use SAM~\cite{ravi2024sam} to generate masks for the detected instances. Finally, we use SigLIP~\cite{zhai2023sigmoid} to evaluate the image-text similarity, taking the region as an image.

To ensure data quality, we apply extensive filtering (\ie, rejection sampling) based on confidence scores from OWL-ViT v2, SAM, and SigLIP image-text similarity. We also ensure we have at most two instances per image, and for images with two instances, these two instances have to be from different classes. The object category names produced by OWL-ViT v2 are then put into a variant of our Describe Anything model, which is trained on data from stage 1 and optimized for self-labeling. This variant generates descriptions with a 50\% probability of incorporating class names during training, as during self-labeling we have a class name as a part of each input. The object category proposals used by OWL-ViT v2 are generated by VILA 1.5~\cite{lin2024vila}.

\noindent\textbf{Detailed localized video captioning.} We annotated 94k regions across 37k videos from SA-V dataset~\cite{ravi2024sam} for detailed localized video captioning, as detailed in~\cref{tab:video_datasets}. Note that each region, also called masklet, indicates an instance across multiple frames in the video. In contrast to the use of SA-1B, where we did not use the masks that come with the dataset, we use the high-quality masklets that come with the videos. We found that many masklets cover parts of an instance, which is not necessarily helpful in describing the whole object as a common use case of our model. Therefore, we performed instance segmentation on the videos with ViTDet~\cite{li2022exploring} + Cascade Mask R-CNN~\cite{cai2019cascade} trained by EVA-02~\cite{eva02} and used voting to match the segmentation masks with the masklets. In this way, we filter out most of the masklets that are parts, since they likely do not correspond to instance masks. The matched masklets carry the class name from the matched instance segmentation mask, which is used in the annotation process to obtain a detailed localized caption for each masklet.

\begin{table*}[ht!]
\setlength\tabcolsep{0pt}
\centering
\begin{tabular*}{\linewidth}{@{\extracolsep{\fill}} l }\toprule
\begin{lstlisting}[style=myverbatim]
You are responsible to write a very descriptive caption to describe the {{category}} in the provided SEGMENTED image. You may leverage the surrounding context of the SEGMENTED image provided in the CROPPED image. 
You must not mention any background in the caption and only describe the {{category}} in the SEGMENTED image! The caption must ONLY contain sufficient details to reconstruct the same {{category}} in the SEGMENTED image but nothing else!
Here are some additional rules you need to follow when describing the {{category}} in the SEGMENTED image:
1. If there are multiple {{category}} in the CROPPED image, focus on the {{category}} in the SEGMENTED image.
2. If the {{category}} in the SEGMENTED image is occluded by other objects, only describe the visible part. DO NOT mention anything that is not directly related to the visible part of {{category}}, such as "A segment of", which part is invisible, etc. For objects with text written on it, describe the object instead of just outputting the text written on it.
Here is the SEGMENTED image that needs caption:
\end{lstlisting} \\\bottomrule
\end{tabular*}
\caption{Our prompt for data annotation in stage 1.}
\label{tab:prompt_for_data_annotation}
\end{table*}

\subsection{Model Training}

We start from off-the-shelf VILA 1.5~\cite{lin2024vila} models that are publicly available on HuggingFace. For image-only training, we fine-tune VILA 1.5 3B model. For joint image-video training, we use VILA 1.5 8B model. We use SigLIP~\cite{zhai2023sigmoid} vision encoder, following VILA 1.5. To prevent catastrophic forgetting and to maintain instruction following capabilities, we mix in ShareGPT-4V~\cite{chen2023sharegpt4v} with our localized image/video captioning dataset collected with our proposed data pipeline. Following the VILA 1.5 training and inference recipe, we treat videos as 8 images concatenated in the sequence.

We closely follow VILA 1.5's recipe of the supervised fine-tuning stage and train all modules, including the vision backbone, the projector, and the LLM. We fine-tune the model for 1 epoch. For the 3B model, we use a batch size of 2048 with a learning rate of 1e-4 on 8 Nvidia A100 GPUs. For the 8B model, we use a batch size of 2048 with a learning rate of 1e-5 on 32 Nvidia A100 GPUs. Both models take less than a day to train. We use a cosine scheduler with a warmup ratio of 0.03. No weight decay is used. For training our model that takes in a class name for self-labeling, we randomly put the class name in the prompt with 50\% probability. For models without prompt augmentation, which is detailed below, we simply use the prompt ``\texttt{Describe the masked region in detail.}'' Following VILA, we always put image tokens in front of the textual tokens. As for the setting for the focal crop, we extend the crop by 1$\times$ the width towards left and right, and 1$\times$ the height towards top and bottom, unless we hit the boundaries of the image, in which case we take the boundaries, \ie $\alpha=3$ and the total area of the crop is enlarged up to $9\times$. If either the height or width is less than 48 pixels, we take 48 pixels for that direction to encode more context for very small regions, since the small regions themselves do not have much useful information.

\noindent\textbf{Prompt Augmentation.} We trained a variant of our model with prompt augmentation to enhance generalization capabilities beyond detailed localized captioning, as analyzed in~\cref{sec:additional_qualitative_results}. For these models, during training, we randomly select one of 15 prompts from a predefined set. These prompts may or may not include a \verb|{prompt_suffix}|. The default prompt suffix is \textit{in detail}. However, we introduce variability by conditioning the prompt on the number of words or sentences in the target caption.

Specifically, with a 20\% probability, we condition the prompt on the number of sentences, using suffixes like \textit{in one sentence} or \textit{in [number of sentences] sentences} (e.g., \emph{in 2 sentences}). If the caption contains only one sentence, we use phrases like \textit{in a sentence} or \textit{in one sentence}.

With another 20\% probability, we condition the prompt on the number of words in the target caption. For captions with a small word count, we use exact numbers (e.g., \textit{in 3 words}). For longer captions (up to 200 words), we may round the word count to the nearest ten and use phrases like \textit{in about 50 words} or \textit{in around 50 words}. If the caption exceeds 200 words, we use the suffix \emph{in more than 200 words}.

The list of prompts that include a \verb|{prompt_suffix}| is as follows:

\begin{enumerate}
\item Describe the masked region \verb|{prompt_suffix}|.
\item Describe the masked area \verb|{prompt_suffix}|.
\item What can you describe about the masked region \verb|{prompt_suffix}|?
\item Can you describe the masked region \verb|{prompt_suffix}|?
\item Provide an explanation of the masked region \verb|{prompt_suffix}|.
\item Depict the masked area \verb|{prompt_suffix}|.
\item Portray the masked area \verb|{prompt_suffix}|.
\item Describe what the masked region looks like \verb|{prompt_suffix}|.
\item Illustrate the masked region \verb|{prompt_suffix}|.
\item How would you explain the masked area \verb|{prompt_suffix}|?
\item What details can you provide about the masked region \verb|{prompt_suffix}|?
\item What does the masked region entail \verb|{prompt_suffix}|?
\item How would you illustrate the masked region \verb|{prompt_suffix}|?
\item How would you depict the masked area \verb|{prompt_suffix}|?
\item How would you portray the masked area \verb|{prompt_suffix}|?
\end{enumerate}

Additionally, we have prompts that inherently request detailed descriptions without requiring a suffix:

\begin{enumerate}
\item Give a detailed description of the masked region.
\item Provide a thorough description of the masked region.
\item Can you explain the details of the masked area?
\item Give a detailed account of the masked region.
\item Describe the masked area comprehensively.
\item Provide an in-depth description of the masked region.
\item Explain the specifics of the masked area.
\item Can you provide a thorough explanation of the masked region?
\item What are the details of the masked area?
\item Provide a comprehensive description of the masked area.
\item What specific details can you provide about the masked region?
\item Can you give an in-depth account of the masked section?
\item What are the main characteristics of the masked region?
\item Give a thorough description of the masked area’s details.
\item Provide detailed information about the masked area.
\end{enumerate}

For prompts without a suffix, we do not condition the generation on the number of words or sentences.

During training, we select prompts based on the \verb|prompt_suffix|:

\begin{itemize}
\item If the \verb|prompt_suffix| is \emph{in detail} (the default option), we may choose from either set of prompts.
\item If the \verb|prompt_suffix| specifies word or sentence counts, we select only from prompts that include \verb|{prompt_suffix}|.
\end{itemize}

This approach introduces variability in the prompts, encouraging the model to generate responses with controls from the prompts in mind, thereby enhancing its generalization and instruction-following capabilities.

\subsection{Inference Setting}
\label{ssec:inference_setting}
Unless otherwise mentioned, our prompt for obtaining detailed localized image descriptions at inference time is the following:

\texttt{Describe the masked region in detail.}

Our prompt for obtaining detailed localized video descriptions at inference time is the following:

\texttt{Given the video in the form of a sequence of frames above, describe the object in the masked region in the video in detail. Focus on appearance, motion, and actions. If the motion involves multiple stages or steps, break down each stage and describe the movements or changes sequentially. Ensure each phase of motion is described clearly, highlighting transitions between actions.}

For Co3Dv2~\cite{reizenstein21co3d} sequences that we treat as videos, we use the following prompt:

\texttt{Describe the masked region in the video in detail. The video consists of multiple views of a stationary object. Focus on the appearance of the object without mentioning any motion or actions.}

\clearpage

{
    \small
    \bibliographystyle{ieeenat_fullname}
    \bibliography{main}
}

\end{document}